\documentclass[acmtog]{acmart}
\pdfoutput=1
\makeatletter                   
\def\mdseries@tt{m}             
\makeatother                    
\usepackage[draft=true]{minted} 
\usepackage{color}
\usepackage{hyperref}           
\hypersetup{
    colorlinks=true,
    linkcolor=blue,
    filecolor=red,      
    urlcolor=magenta,
    breaklinks=true,            
}
\usepackage{breakurl}           

\usepackage{bm}
\usepackage{multirow}

\newcommand{\R}{\mathbb{R}}
\usepackage{booktabs}
\usepackage{tabularx}
\usepackage{lscape}
\setcopyright{none}
\begin{document}
\sloppy

%

\title{3D Morphable Face Models - Past, Present and Future} 


\author{Bernhard Egger}
\affiliation{\institution{Massachusetts Institute of Technology, USA}}
\email{egger@mit.edu}
\author{William A. P. Smith}
\affiliation{\institution{University of York, UK}}
\email{william.smith@york.ac.uk}
\author{Ayush Tewari}
\affiliation{\institution{Max Planck Institute for Informatics \& Saarland Informatics Campus, Germany}}
\email{atewari@mpi-inf.mpg.de}
\author{Stefanie Wuhrer}
\affiliation{\institution{Univ. Grenoble Alpes, Inria, CNRS, Grenoble INP$^*$\thanks{$^*$Institute of Engineering Univ. Grenoble Alpes}, LJK, 38000 Grenoble, France}}
\email{stefanie.wuhrer@inria.fr}
\author{Michael Zollhoefer}
\affiliation{\institution{Stanford University, USA}}
\email{michael@zollhoefer.com}
\author{Thabo Beeler}
\affiliation{\institution{Disney Research|Studios, Switzerland}}
\email{thabo.beeler@disneyresearch.com}
\author{Florian Bernard}
\affiliation{\institution{Max Planck Institute for Informatics \& Saarland Informatics Campus, Germany}}
\email{fbernard@mpi-inf.mpg.de}
\author{Timo Bolkart}
\affiliation{\institution{Max Planck Institute for Intelligent Systems, Germany}}
\email{tbolkart@tuebingen.mpg.de}
\author{Adam Kortylewski}
\affiliation{\institution{Johns Hopkins University, USA}}
\email{akortyl1@jhu.edu}
\author{Sami Romdhani}
\affiliation{\institution{IDEMIA, France}}
\email{sami.romdhani@idemia.com}
\author{Christian Theobalt}
\affiliation{\institution{Max Planck Institute for Informatics \& Saarland Informatics Campus, Germany}}
\email{theobalt@mpi-inf.mpg.de}
\author{Volker Blanz}
\affiliation{\institution{University of Siegen, Germany}}
\email{blanz@informatik.uni-siegen.de }
\author{Thomas Vetter}
\affiliation{\institution{University of Basel, Switzerland}}
\email{thomas.vetter@unibas.ch}
\renewcommand{\shortauthors}{B. Egger et al.}
\begin{abstract}

In this paper, we provide a detailed survey of 3D Morphable Face Models over the 20 years since they were first proposed. The challenges in building and applying these models, namely capture, modeling, image formation, and image analysis, are still active research topics, and we review the state-of-the-art in each of these areas. We also look ahead, identifying unsolved challenges, proposing directions for future research and highlighting the broad range of current and future applications.
\end{abstract}

\begin{teaserfigure}
  \includegraphics[width=\textwidth]{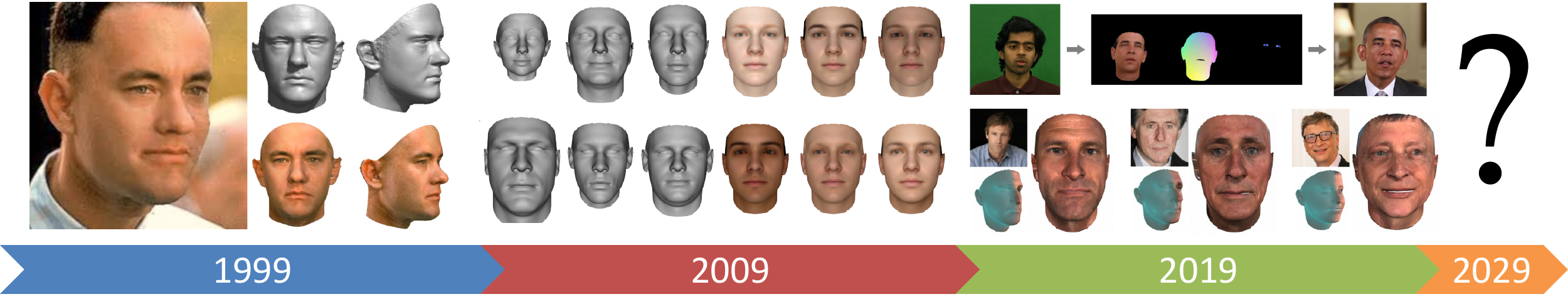}
 \caption{20 years of 3D Morphable Models. Fitting results from the original paper \cite{BlanzVetter1999}, the first publicly available Morphable Model \cite{paysan20093d}, and state-of-the-art facial re-enactment results \cite{kim2018DeepVideo} and GAN-based models \cite{gecer2019ganfit}.}
  \Description{}
  \label{fig:teaser}
\end{teaserfigure}

\maketitle
\thispagestyle{empty}

\section{Introduction}

It is 20 years since 3D Morphable Face Models were first presented at SIGGRAPH '99. They were proposed as a general face representation and a principled approach to image analysis. \citet{BlanzVetter1999} introduced and tackled many subsidiary problems and the results were considered groundbreaking. The impact of the original paper has been long term, recognized by an impact paper award, and the approach and applications are accessible to a wide audience (the original supplementary video was one of the most popular videos in the early days of YouTube). However, the approach is not just of historical interest. In the past two years, 3D Morphable Face Models have been re-discovered in the context of deep learning and are incorporated into many state-of-the-art solutions for face analysis. This survey aims to build a starting point for researchers new to the topic, act as a reference guide for the community around 3D Morphable Models and to introduce exciting open research questions.

\subsection{Definition}
A 3D Morphable Face Model is a generative model for face shape and appearance that is based on two key ideas: First, all faces are in dense point-to-point correspondence, which is usually established on a set of example faces in a registration procedure and then maintained throughout any further processing steps. Due to this correspondence,  linear combinations of faces may be defined in a meaningful way, producing morphologically realistic faces (\textit{morphs}). The second idea is to separate facial shape and color and to disentangle these from external factors such as illumination and camera parameters. The Morphable Model may involve a statistical model of the distribution of faces, which was a principal component analysis in the original work \cite{BlanzVetter1999} and has included other learning techniques in subsequent work.
\begin{figure}
    \centering
    \includegraphics[width=\linewidth]{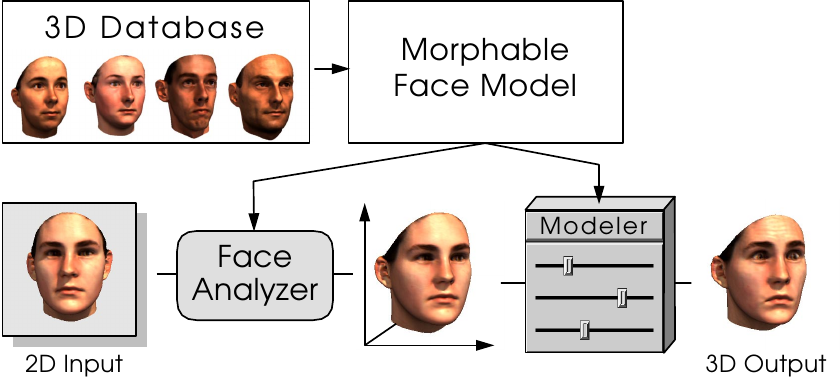}
    \caption{The visual abstract of the seminal work by \citet{BlanzVetter1999}. It proposes a statistical model for faces to perform 3D reconstruction from 2D images and a parametric face space which enables controlled manipulation.}
    \label{fig:momo99}
\end{figure}

\subsection{History}
The initial research question behind the idea of 3D Morphable Models (3DMM) was how a visual system, biological or artificial, can cope with the high variety of images that a single class of objects can generate, and how objects are represented to solve vision tasks.  
The leading assumption for the development of 3DMMs was that prior knowledge about object classes plays an important role in vision and helps to solve otherwise ill-posed problems. 3DMMs are designed to capture such prior knowledge, and they are learned automatically from a set of examples. The representation is general, so it may be applied to different objects and tasks.

Representations of faces and the task of face recognition have been in the focus of vision research for a long time. An important and very influential paradigm shift in this field was the Eigenfaces approach by \citet{sirovich1987low} and \citet{turk1991eigenfaces}, which learned an explicit face representation from examples and operated
entirely on grey-levels in the image domain. 
Eigenfaces treated images of faces as a vector space and performed a principal component analysis, with the eigenvectors representing the main modes of variation in that space. The drawback of Eigenfaces was not only that it was limited to a fixed pose and illumination, but that it had no effective representation of shape differences: When the coefficients in
linear combinations of eigenvectors are changed continuously, structures will fade in and out, rather than shift along the image plane. As a consequence, the model fails to find a single parameter for, say, the distances between the eyes. 
 The Eigenfaces approach was also extended to 3D face surfaces by \citet{atick1996statistical} to model shading variations in faces, yet with essentially the same limitation.

Several research groups proceeded by adding an Eigendecomposition of 2D shape variations between individual faces. This provided both an explicit shape model, and - after warping the images - an aligned Eigenface model without blurring and ghosting artifacts. While in the original Eigenface approach, the images were only aligned by a single point (e.g., the tip of the nose), the new methods established correspondence on significantly more points. Landmark-based face warping for image analysis was introduced by \citet{craw1991parameterising}. Using approximately 200 landmarks, the first statistical shape model was proposed in Active Shape Models  \cite{cootes1995asm}. While this model used shape only, Active Appearance Models \cite{Cootesetal98} proposed a combination of shape and appearance that turned out to be very successful and influential. Other groups computed dense pixel-wise image correspondences with optic-flow algorithms for modeling the facial shape variations \cite{hallinan1999, jones1998multidimensional}. In all these correspondence-based approaches, images are warped to a common template, and the appearance variation is then performed in the same way as the original Eigenfaces, but on the shape-normalized images. The shape model, on the other hand, provides a powerful and compact representation of shape differences by shifting pixels in the image plane. However, compared to the simple linear projection in Eigenfaces, the image analysis task is transformed into a more challenging nonlinear model fitting problem.

These 2D models were efficient to cover the shape variation for a fixed pose and illumination setting. The framework was extended to variations across pose by \citet{vetter1997linear} and to other object classes, such as images of cars \cite{jones1998multidimensional}. All this groundwork demonstrated that a separation of shape and texture information in images can model the variation of faces. On the other hand, the price to pay for taking pose and illumination variations into account was high: eventually, it would require many separate models, each limited to a small range of poses and illuminations. In contrast, the progress of 3D Computer Graphics in the 1990s demonstrated that variations in pose and illumination are easy to simulate, including self-occlusion and shadowing. Adapting methods from graphics to face modeling and computer vision led to the new face representation in 3DMMs and the idea of using analysis-by-synthesis to map between the 3D and 2D domain. Those were the two key contributions in the first paper on 3DMMs  \cite{BlanzVetter1999}, compare Figure~\ref{fig:momo99}. The name Morphable Model was derived from their 2D counterpart \cite{jones1998multidimensional}, and in fact, Jones and Poggio strongly influenced the ideas that led to 3DMMs.

3DMMs and 2D Morphable Models rely on dense correspondence, rather than only a set of facial feature points. In the original work, this was established by an optical flow algorithm for image registration. The image synthesis algorithm used a standard rendering model with perspective projection, ambient and directional lighting, and a Phong model of surface reflectance that includes a specular component. However, in analysis-by-synthesis, this approach comes at a computational price because shape-camera \cite{smith2016perspective} and illumination-albedo \cite{egger17phd} ambiguities lead to a hard ill-posed optimization problem. Moreover, the optimization is costly and is prone to end in unwanted local optima. Just as it is already dramatically more complicated to fit an Active Appearance Model to a 2D image, compared to the simple projection needed for Eigenfaces, the complexity of 3DMM fitting raises additional problems which have remained challenging to researchers after 20 years of development.

At the time the initial 3DMM was developed, image-based models were dominating computer vision and even animation \cite{ezzat2002}, and they were rather elaborate at that time. It was a key decision to take the best of both the 2D and the 3D world, by using 3D models to manipulate existing images on the one hand, and applying 2D algorithms to 3D surfaces: Unlike mesh-based algorithms, the original 3DMM used optical flow, multi-resolution approaches and interpolation algorithms on parameterized surfaces of faces. With the initial face scanner delivering surfaces in a two-dimensional cylinder parameterization, all those steps were performed in 2D, and most of the methods involved were replaced with their 3D equivalent only many years later.
It is interesting to see that after a development towards 3D, the computer vision community came back to 2D representations by using deep learning, and now evolves again to 3D, e.g.,~by integrating 3DMMs.

Over the past years, 3DMMs were applied beyond faces. Models were built for the surface of the human body \cite{allen2003space, anguelov2005scape, SMPL:2015} and for other specific parts of the body like ears \cite{dai2018data} and hands \cite{khamis2015learning}, animals \cite{Zuffi:CVPR:2018, Sun_2020_WACV} and even cars \cite{shelton2000morphable}. In this survey, we focus on 3DMMs to model the human face, though many of the techniques and challenges are the same across different object classes.

The 3DMM was developed in a time where algorithms and data were rarely shared across researchers and institutions. 10 years later the first publicly available 3DMM was released \cite{paysan20093d} and in the last 10 years, all individual data and algorithmic components needed to build and use 3DMMs were released by various researchers. We collected a list of all available resources and will further maintain it \cite{OverviewGithub2019}. 

The 3DMM was built as a general representation for faces, not just aiming at one specific task. Even though the model is outperformed for some very specific applications such as face recognition, it is unique in its generality across different tasks and applications.

\subsection{Organization}
There is a recent state of the art report on monocular 3D reconstruction, tracking and applications \cite{ZollhoeferSTAR2018}. This focuses on the most recent advances, particularly related to the specific task of tracking and reconstruction. In contrast, in this paper, we instead focus on the 3DMM, all involved methods, and reflect the major contributions over the past 20 years while at the same time highlighting challenges and future directions.

This survey is organized from building to applying a 3DMM. We start with Section \ref{sec:capture} where we present methods to acquire 3D facial data for model building. We then describe in Section \ref{sec:modelling} the various approaches to model the 3D shape and facial appearance. In Section \ref{sec:synthesis} we discuss the methods to generate a 2D image from our 3D model using computer graphics. Our Section \ref{sec:AbyS} surveys the major application of 3DMMs, namely the reconstruction of a 3D face from a 2D image. Section \ref{sec:deeplearning} summarizes the impact of 3DMMs in the recent advances in the field of deep learning and how deep learning can be used to improve the modeling and analysis. Section \ref{sec:applications} summarizes the various applications where 3DMMs were used in the past 20 years. Every Section summarizes the major challenges the authors see regarding the current limitations of 3DMM. We also collect challenges that are shared across multiple Sections in Section \ref{sec:perspective}, where we also venture an outlook on what we expect to see in the next 10 to 20 years and how the 3DMM will keep impacting how faces are represented.

\section{Face Capture}
\label{sec:capture}


\begin{figure*}
    \centering
    \bgroup
    \setlength\tabcolsep{0.05cm}
    \begin{tabular}{cc}
    \begin{tabular}{c}
    \includegraphics[height=2.5cm]{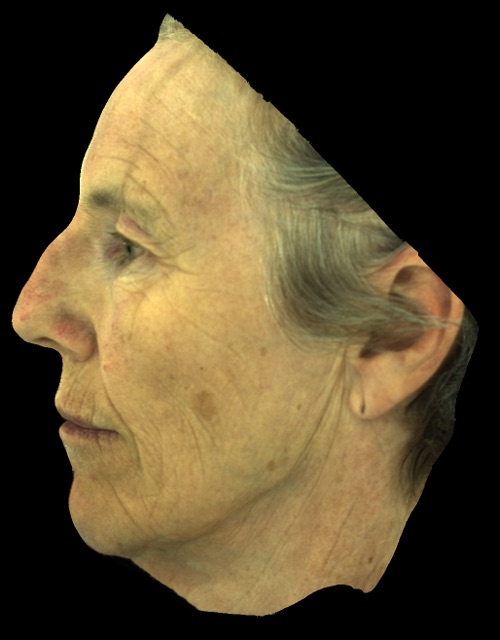}
    \includegraphics[height=2.5cm]{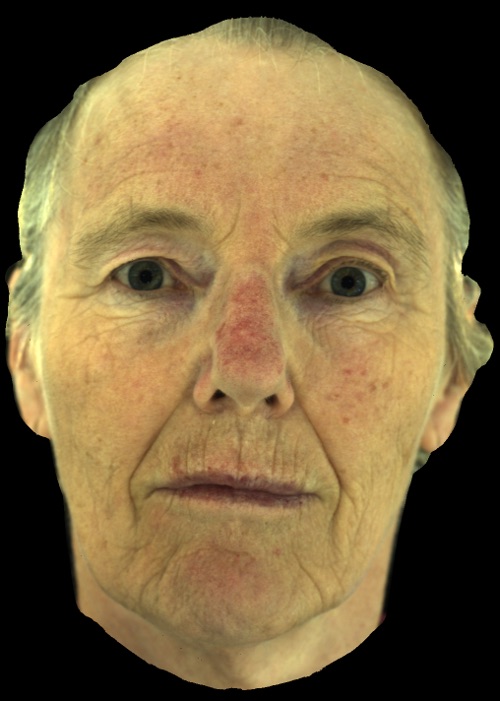}
    \includegraphics[height=2.5cm]{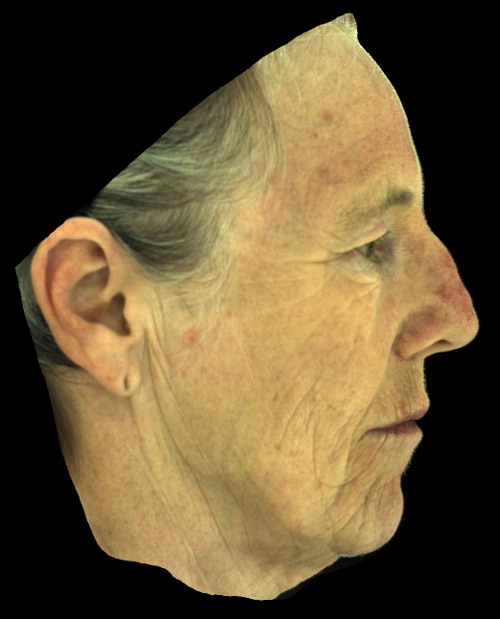} \\
    (a) Diffuse albedo\\
    \includegraphics[height=2.5cm]{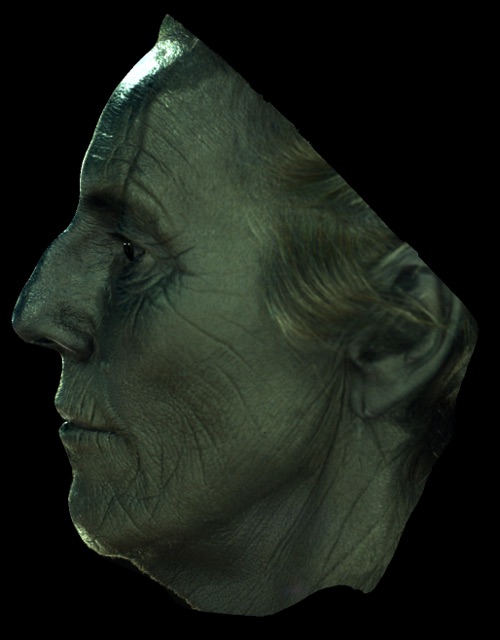}
    \includegraphics[height=2.5cm]{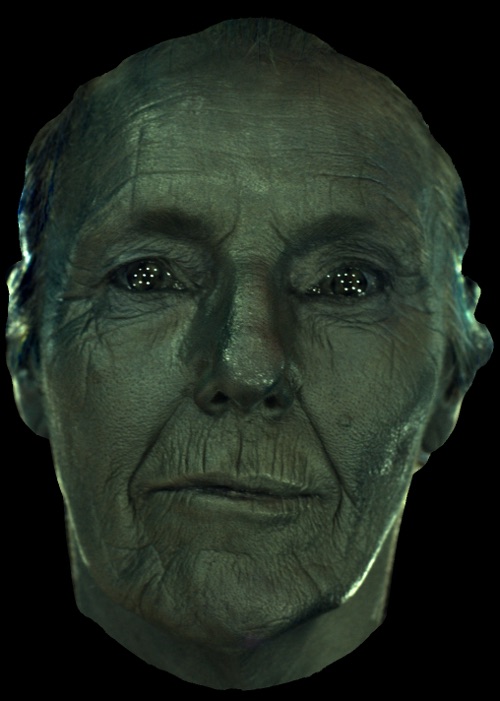}
    \includegraphics[height=2.5cm]{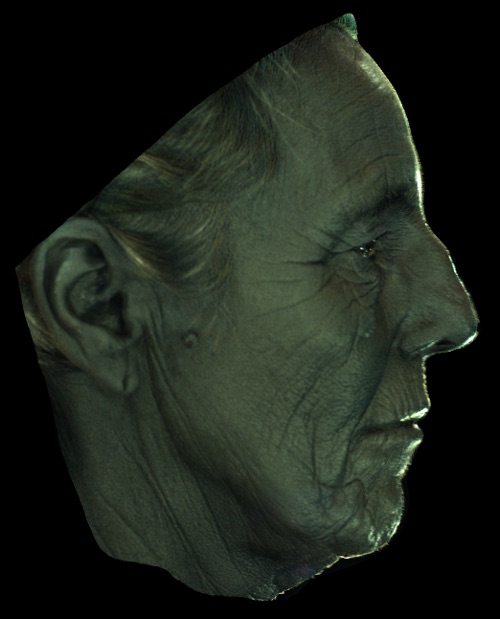}\\
    (b) Specular albedo\\
    \end{tabular}
    \begin{tabular}{ccc}
    \includegraphics[height=5.5cm]{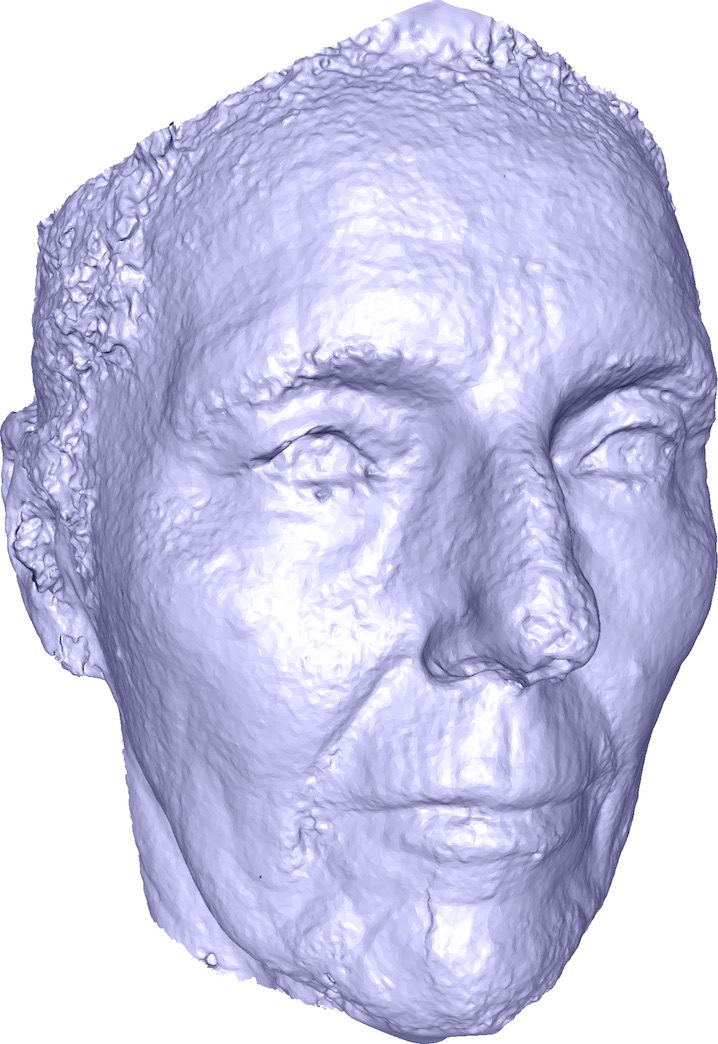} &  
    \includegraphics[height=5.5cm,trim=11px 0px 22px 0px]{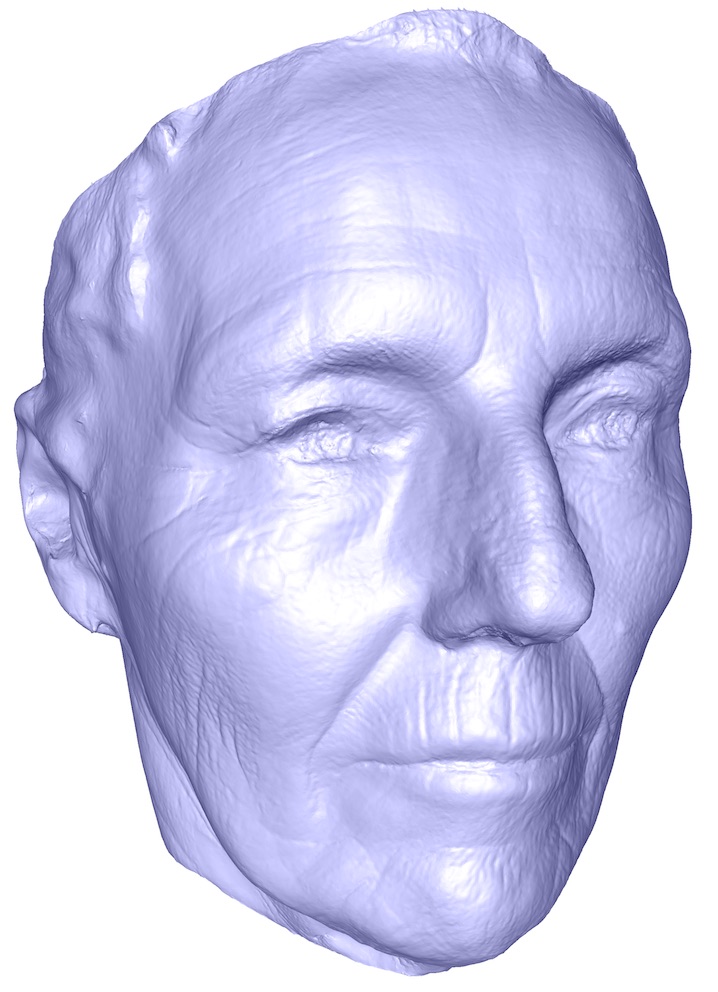} & 
    \includegraphics[height=5.5cm,trim=11px 0px 22px 0px]{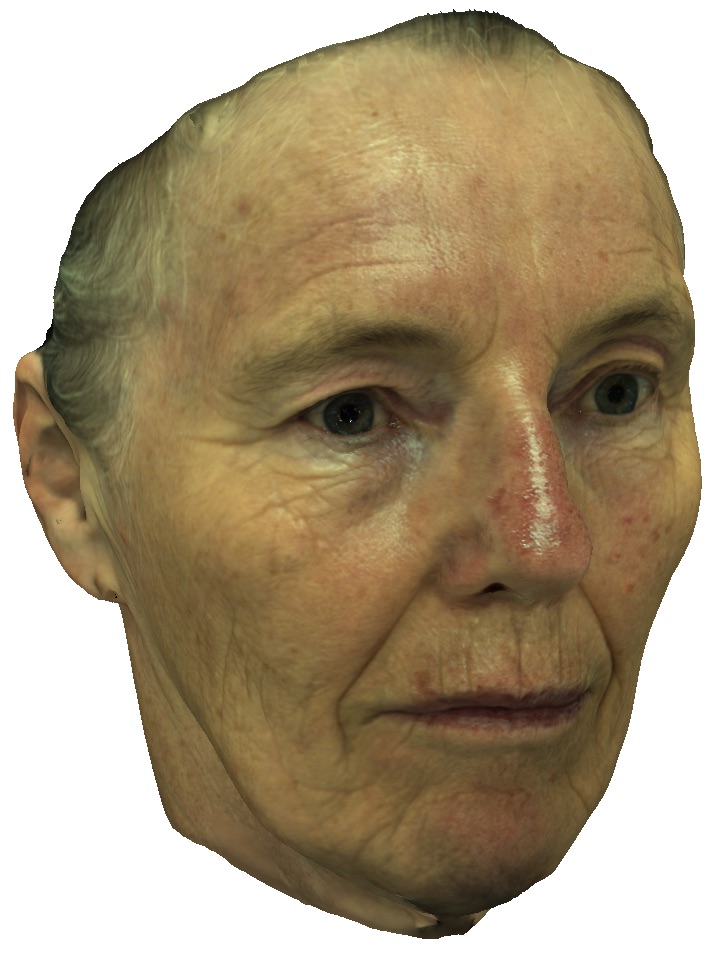} \\
    (c) MVS geometry & (d) Hybrid geometry & (e) Rendering \\
    \end{tabular}
    \end{tabular}
    \egroup
    \caption{Capture of intrinsic face properties using a hybrid geometric/photometric method \cite{seck2016ear, smith2020albedo}. Multi view stereo (MVS) is used to reconstruct a coarse mesh (c). A photometric light stage \cite{ma2007rapid} is used to capture diffuse and specular albedo maps (a,b) and surface normals that are merged with the MVS mesh to produce a mesh with fine surface detail (d). Together, these can be used to synthesize highly realistic images of the face (e).}
    \label{fig:capture}
\end{figure*}

The key ingredient to any 3DMM is a representative set of 3D shapes, usually coupled with corresponding appearance data. The typical way to construct such a sample pool is by acquiring data from the real world. In this section, we give a 
brief
overview of different approaches that have been used to acquire facial data as well as data of facial parts. As we are concerned with the creation of input datasets for 3DMMs, we limit the discussion to acquisition under controlled conditions, as opposed to the more challenging in-the-wild setting. Note that controlled 3D face capture may not always be necessary. There have been attempts to learn 3DMMs directly from images \cite{cashman2012shape} and state-of-the-art deep learning-based methods simultaneously learn a 3DMM and regression-based fitting from 2D training data (see Section \ref{sec:3Dfrom2D}). 
In this section we begin by covering shape acquisition methods in Section \ref{subsec:shapeac} including geometric, photometric and hybrid methods. Sections \ref{subsec:appcapture}, \ref{subsec:faceparts} and \ref{subsec:dyncap} describe methods for capture of appearance, face parts and dynamics respectively. Section \ref{sec-datasets} lists publicly available 3D face datasets that could be exploited for building 3DMMs. Finally, we consider open challenges related to face capture in Section \ref{subsec:capchallenges}.

\subsection{Shape Acquisition}\label{subsec:shapeac}
The three-dimensional shape is arguably the most important ingredient to a 3DMM. The issue of shape representation has not been widely considered in the context of 3DMMs. By far the most commonly used representation is a triangle mesh. Rare exceptions include cylindrical \cite{atick1996statistical} and orthographic \cite{dovgard2004statistical} depth maps (though these representations do not permit meaningful dense correspondence), per-vertex surface normals \cite{aldrian2012inverse}, and, more recently, volumetric orientation fields \cite{saito20183d} and signed distance functions \cite{park2019deepsdf}. Using a triangle mesh representation, dense correspondence requires that all samples exhibit the same topology and that the vertices encode the same semantic point on all samples. Establishing correspondence across the samples is a challenging topic in itself, discussed in Section \ref{subsec:correspondence}. In this section, we focus on the acquisition of raw 3D data, before establishing correspondence.


\subsubsection{Geometric methods}

Geometric methods estimate directly the 3D coordinates of a shape either by observing the same surface point from two or more viewpoints (in which case the challenge is identifying corresponding points between images) or by observing a projected pattern (in which case the challenge is identifying the correspondence between the known pattern and an image of its projection). Methods can either be considered active, i.e.~they emit light or other signals into the scene, or passive. Laser scanners, Time-of-Flight sensors, and Structured Light systems are active systems, where multi-view photogrammetry is a passive alternative. Active multi-view photogrammetry may be considered a hybrid active/passive approach, as it relies on passive photogrammetry to reconstruct the shape, but augments the object with a well-defined texture projection that benefits the reconstruction \cite{zhang2004spacetime}. Unlike structured light, the origin of the light does not matter as the projected texture is solely meant to augment the texture used for multi-view stereo matching. This type of technology is used by the Intel\textsuperscript{\textregistered} RealSense{\texttrademark} D435 camera for example\footnote{https://www.intelrealsense.com/depth-camera-d435/}. In the early days of 3DMMs, active systems were the only real option to acquire 3D shapes at a reasonable quality. The original paper of \citet{BlanzVetter1999} relied on laser scanning \cite{levoy2000digital}, where the face is rasterized via one or more laser-beams. The laser beam illuminates the face surface at a point and using the known camera/laser arrangement the 3D position of this point may be triangulated. The biggest drawback of laser scanners is the acquisition time, as only very few samples are gathered at any given time -- even at very high frame rates, such systems require the subjects to sit still for several seconds. 

Structured light scanners \cite{Geng11structured} overcome this limitation to some extent by injecting not only a few beams but leveraging projectors that offer millions of them. The challenge here is to identify which beam is illuminating the object at a given point. This is addressed by structuring the projected light in a way that allows to clearly identify the origin of any ray. The simplest approach is binary encoding, which projects black and white patterns assigning a unique binary code to each pixel. The required number of patterns is still quite substantial, for VGA resolution one needs 19 distinct patterns and for 4K resolution 23 patterns, and hence this approach is most suited for capturing static objects. However, technical improvements have begun to make these approaches viable for dynamic capture of faces. The Intel\textsuperscript{\textregistered} RealSense{\texttrademark} SR300 uses only 9 binary patterns to obtain VGA, while the most recent RealSense depth camera produces VGA resolution at 60 depth FPS with a scanning laser technology. Other more complex structured light methods have been proposed, such as gray codes or (colored) fringe patterns, which can reduce the number of required frames further, in extreme cases even to a single frame. A very popular commercial system that was used to create face datasets (\cite{Cao2014_FaceWarehouse}) and that employs structured light is the first generation Kinect sensor\footnote{https://en.wikipedia.org/wiki/Kinect\#Kinect\_for\_Xbox\_360\_(2010)}. The device employs a structured dot pattern, which allows reconstructing depth from a single frame by sacrificing spatial resolution. Resolution may be improved by accumulating several frames \cite{newcombe2011kinectfusion}. With the increased resolution and quality of consumer cameras, passive systems have become the method of choice in most cases, since they are simpler to assemble and operate and off-the-shelf photogrammetry software solutions, both commercial such as Agisoft\footnote{https://www.agisoft.com} or RealityCapture\footnote{https://www.capturingreality.com}, as well as open-source solutions such as Meshroom\footnote{https://alicevision.org/}, provide very good results on human faces. Also, complete systems can be purchased that come with both hard- and software\footnote{https://www.canfieldsci.com/imaging-systems/vectra-m3-3d-imaging-system/}\footnote{http://www.di4d.com}\footnote{http://www.3dmd.com/}. These methods typically do not require the aggregation of information over time and hence offer themselves for single-shot acquisition \cite{beeler2010high} as well as full-frame rate performance capture \cite{furukawa2009dense,bradley2010high,Beeler2011}. A potential disadvantage of the aforementioned systems is their form factor since they all require at least some separation between the different participating components, i.e. the cameras or lights, often referred to as the baseline. An alternative which becomes more and more viable due to the push of the mobile industry are time-of-flight sensors, where the elements can be located close to each other. The second-generation Kinect sensor\footnote{https://en.wikipedia.org/wiki/Kinect\#Kinect\_for\_Xbox\_One\_(2013)} belongs to this family, as well as many depth sensors that are shipped with modern mobile phones. A challenge that time-of-flight sensors share with most of the active systems is that color information has to be acquired separately and is not intrinsically aligned with the 3D data, which is another advantage of passive setups. 

\subsubsection{Photometric methods}

Photometric methods typically estimate surface orientation, from which the 3D shape may be recovered via integration. The challenge here is to select models that accurately capture reflectance properties of the surface and obtaining sufficient measurements that the inversion of these models is well-posed. Compared to geometric methods, photometric methods typically offer higher shape detail and do not rely on the presence of matchable features (so are applicable to smooth, featureless surfaces) but often suffer from low-frequency bias in the reconstructed positions caused by modeling errors in reflectance and illumination. 
Photometric stereo \cite{ackermann2015survey} estimates the surface normal at each pixel by observing a scene from a fixed position under at least three different illumination conditions, which can be spectrally multiplexed \cite{hernandez2007non} in order to reduce the number of frames required. Early work assumed known lighting directions and perfectly diffuse reflectance. When illumination is uncalibrated and a more suitable glossy reflectance model is used, generic face priors can be used to resolve the resulting ambiguity \cite{georghiades2003incorporating}. Typically more lighting conditions are used to increase robustness and coverage, such as four \cite{zafeiriou2013face} or even nine \cite{gotardo2015ICCV}. Gradient-based illumination takes the number of conditions to the extreme, by illuminating the subject not with discrete individual point lights, but by an ideally continuous, omnidirectional incident illumination gradient. An advantage of this set up is that hard light source occlusions (cast shadows) are replaced by soft partial occlusions of the illuminating hemisphere (ambient occlusion). In practice, the omnidirectional illumination is realized via a light-stage \cite{debevec2000acquiring}, which discretizes the gradient with a large number (several hundred) of light sources. The original work of \citet{ma2007rapid} suggests the use of four distinct gradients, which has later been extended using complementary gradients \cite{wilson2010temporal}. Again, variants of temporal, spectral and polarization multiplexing have been proposed to reduce the number of required conditions.

\subsubsection{Hybrid methods}

Hybrid methods combine the strength of geometric and photometric methods, specifically, they reduce the low-frequency bias typically present in photometric methods and increase the high-frequency details when compared to geometric methods. \citet{nehab2005efficiently} propose a method for merging the low frequencies of positional information and the high frequencies of surface normals. The method is particularly efficient, involving only the solution of a sparse linear system of equations, and has been used in the context of 3DMM fitting \cite{patel2012driving}. Various combinations of geometric and photometric methods have been considered. For example, \citet{zivanov2009facial} combine structured light with photometric stereo, \citet{ma2007rapid} combine structured light with gradient-based illumination, \citet{ghosh2011multiview} combine multi-view stereo with gradient-based illumination, and \citet{beeler2010high} combine passive multi-view photogrammetry with shape-from-shading. Figure \ref{fig:capture}(d) shows the output of a hybrid method in which photometric surface normals are merged with a multi-view stereo mesh. 

\subsection{Appearance Capture}\label{subsec:appcapture}

In addition to shape, appearance is also required for many 3DMM tasks, such as synthesizing images (see Section \ref{sec:synthesis}) and inverse rendering (see Section \ref{sec:AbyS}). Unlike shapes, which are almost exclusively represented as triangular meshes, appearance representation varies substantially. While in theory, every vertex of the mesh could have an associated appearance property, typically shapes are parameterized to the 2D domain and textures are used to store appearance properties. Appearance can be as simple as backprojecting the color of the images onto the shapes, which causes shading effects to be baked in. Self-occlusion, in particular when only a single viewpoint is available, results in missing data in the occluded areas, which must be hallucinated somehow. \citet{booth20183d} use 3DMM fits to in-the-wild images and Principal Component Pursuit with missing values to complete the unobserved texture. They build their appearance model directly on the sampled textures. Such a simplistic approach, however, does not allow intrinsic face appearance properties to be separated from shading/shadowing (and hence illumination/geometry). A partial solution to this problem is to control illumination conditions during capture, for example by using multiple light sources to create approximately ambient lighting. Note that a truly Lambertian convex surface observed under truly ambient light gives exactly the albedo \cite{lee2005acquiring}. The appearance models in the most popular 3DMMs \cite{paysan20093d,booth2018large,LYHM2017} use this approach, combining images from multiple cameras to provide full coverage of the face with diffuse lighting to approximate albedo. A better approach is to explicitly separate shading from skin color, often referred to as intrinsic decomposition. This allows relighting of the face under novel incident illumination conditions and a 3DMM built on such data truly models intrinsic characteristics of the face. 
Several approaches have been presented over the years to acquire reflectance data suited for parametric rendering, measuring surface reflectance \cite{marschner1999image} and even subsurface scattering properties \cite{ghosh2008practical}. The polarised spherical illumination environment used by \citet{ma2007rapid} enables diffuse albedo to be captured in a single shot and specular albedo in two images (see Figure \ref{fig:capture}(a) and (b)). While such approaches have predominately used active setups, recently capture under passive conditions has been demonstrated \cite{gotardo2018practical}.

\subsection{Face part specific methods}\label{subsec:faceparts}

Certain parts of the human face require more targeted acquisition methods and devices since they do not conform with the assumptions typically made by abovementioned approaches. For example, the frontmost part of the eye, the cornea, is for obvious reasons fully transparent and distorts the appearance of the underlying iris due to refraction. \citet{berard2014high} leverage a combination of several specialized algorithms, including shape-from-specularity, in order to reconstruct all visible components of the eye. Another challenging example are teeth \cite{wu2016teeth}, which exhibit extremely challenging appearance \cite{velinov2018appearance}. Hair violates the common assumption that the reconstructed shape is a smooth continuous surface, and requires specialized approaches that estimate hair fibers \cite{beeler2012coupled}, hair strands \cite{luo2013structure,hu2014robust} and braiding \cite{hu2014capturing}, or even encode hair as a surface \cite{echevarria2014capturing} for manufacturing. While most hair acquisition focuses on static reconstruction, some do capture hair in motion \cite{xu2014dynamic} or estimate physical properties for hair simulation \cite{hu2017simulation}. Especially challenging is the acquisition of partially or completely hidden properties, such a the tongue \cite{hewer2018multilinear}, the skull \cite{achenbach2018multilinear,beeler2014rigid}, or the jaw \cite{zoss2018empirical, zoss2019accurate}, where oftentimes specialized imaging systems are required, such as Computer Tomography (CT), Magnetic Resonance Imaging (MRI), or Electromagnetic Articulography (EMA). Lastly, even skin itself requires specialized treatment in some areas, such as lips \cite{garrido2016corrective} or eyelids \cite{bermano2015detailed}, where the local appearance and deformation exceed the capabilities of the more generic methods.

\subsection{Dynamic capture}\label{subsec:dyncap}
Historically, 3DMMs have been mostly concerned with static shapes, for example with a set of neutral shapes from different individuals or with a discrete set of expressions per individual, neglecting how the face transitions between expressions. Most capture systems used to build 3DMMs were hence static systems, focused on capturing individual shapes rather than full performances. As the field begins to integrate more temporal information into the models, the need for dynamic capture systems will rise. Active systems have been considered, both geometric \cite{zhang2004spacetime} and photometric \cite{wilson2010temporal}. However, passive systems \cite{bradley2010high,Beeler2011} are currently the technologies of choice, since they do not require temporal multiplexing and still deliver high-quality shapes, and more recently even per-frame reflectance data \cite{gotardo2018practical}. A beneficial side-effect of such technologies is that they often provide shapes that are already in correspondence, removing the need to establish correspondence in a post-processing step (Section~\ref{subsec:correspondence}), and making them attractive solutions even when only a discrete set of shapes is desired. Available commercial solutions include Di4D\footnote{http://www.di4d.com/}, 3dMD\footnote{http://www.3dmd.com/}, or the Medusa system\footnote{https://studios.disneyresearch.com/medusa/}.

\subsection{Publicly available face datasets}
\label{sec-datasets}
A relatively large number of publicly available datasets exist that could be leveraged in the construction of 3DMMs, though many have never been used for this purpose. We believe there is not broad awareness of the range of 3D datasets available and so collect them together in Table \ref{table:datasets}. We hope that this will encourage work that seeks to exploit multiple datasets for 3DMM building.
\begin{table*}[!t]
\small{
\begin{tabular}{m{0.25\linewidth} m{0.22\linewidth} m{0.125\linewidth} m{0.22\linewidth} m{0.08\linewidth}} 
\toprule
\textbf{dataset} & \textbf{format and resolution} & \textbf{coverage} & \textbf{no. samples} & \textbf{scanner} \\ \midrule

Spacetime faces \cite{zhang2004spacetime} & triangle mesh (23k vertices, consistent topology) & inner face only & 1 individual $\times$ 384 frame dynamic sequence & structured light \\ \hline
CASIA 3D Face Database \cite{casia3D}& 640$\times$480 depth map and texture image& face, neck, sometimes ears & 123 individuals $\times$ 37-38 scans (expression, pose, illumination) & Minolta Vivid910\\ \hline
BU-3DFE \cite{BU-3DFE_2006} & triangle mesh (20k-35k triangles), two texture images (1,300 $\times$ 900) & face, neck, sometimes ears & 100 individuals $\times$ 25 expressions & 3dMD \\ \hline
BU-4DFE \cite{BU-4DFE_2008} & triangle mesh (35k vertices), texture image (1,040 $\times$ 1,329) & face, neck, sometimes ears & 101 individuals $\times$ six 100 frame expression sequences & Dimensional Imaging \\ \hline
Bosphorus~\cite{Bosphorus_2008}& $1,600\times 1,200$ depth map and texture image & inner face only & 105 individuals $\times$ up to 35 expressions per subject $+$ 13 poses &  Inspeck Mega Capturor II \\ \hline
York 3D Face Database \newline \cite{heseltine2008three}& depth map containing 5k-6k points, texture image & inner face only & 350 individuals $\times$ 15 expressions & projected pattern stereo \\ \hline
B3D(AC)\^{}2~\cite{B3D(AC)2010} & raw scan: triangle mesh (55k vertices), 780 $\times$ 580 texture image; processed: triangle mesh (23k vertices, consistent topology), 1,024 $\times$ 768 UV texture map & inner face only & 14 individuals $\times$ around 80 dynamic sequences (speech-4D) & structured light stereo \\ \hline
Florence 3D Faces \cite{Bagdanov2011} & triangle mesh (60k-80k triangles), 4 MPixel texture, additonal 2D HD video & face, neck, sometimes ears & 53 individuals & 3dMD \\ \hline
D3DFACS~\cite{D3DFACS2011} & triangle mesh (30k vertices), 1,024 $\times$ 1,280 UV texture map & face, neck, sometimes ears & 10 individuals $\times$ around 52 dynamic sequences, FACS coded & 3dMD  \\ \hline
3DRFE \cite{stratou2011effect} & triangle mesh (1.2M vertices), 1,296 $\times$ 1,944 diffuse and specular albedo maps and hybrid normal maps & inner face, neck & 23 individuals $\times$ 15 expressions & light stage \\ \hline
Hi4D-ADSIP~\cite{Hi4D-ADSIP2012} & triangle mesh (20k vertices), texture image & inner face only & 80 individuals $\times$ around 42 dynamic sequences & Dimensional Imaging \\ \hline
BP4D-Spontaneous \cite{BP4D-Spontaneous2014} & triangle mesh (30k-50k vertices), texture image (1,040 $\times$ 1,329) & face, neck, sometimes ears & 41 individuals $\times$ eight one minute dynamic sequences & Dimensional Imaging \\ \hline
3D Dynamic Database for Unconstrained Face Recognition \newline \cite{Alashkar2014} & 3.5k vertices for dynamic, 50k vertices for static, texture image & inner face only & 58 individuals $\times$ one static scan + seven dynamic sequences & Artec \\ \hline
FaceWarehouse~\cite{Cao2014_FaceWarehouse} & raw: 640 $\times$ 480 RGBD; processed: triangle mesh (11k vertices, consistent topology) & & 150 individuals $\times$ 20 expressions & Microsoft Kinect \\ \hline
MMSE~\cite{MMSE2016} & triangle mesh (30k-50k vertices), 1,040 $\times$ 1,392 texture image & inner face only & 140 individuals $\times$ four dynamic sequences & Dimensional Imaging \\ \hline
Headspace~\cite{LYHM2017} & triangle mesh (180k vertices), 2,973 $\times$ 3,055 UV texture map & full head including face, neck, ears & 1,519 individuals & 3dMD \\ \hline
4DFAB~\cite{4DFAB2018} & triangle mesh (60k-75k vertices), UV texture map & face, neck and ears & 180 individuals $\times$ 4k-16k frames of dynamic sequences & Dimensional Imaging \\ \hline
CoMA~\cite{CoMA2018} & triangle mesh (80k-140k vertices), texture images (avg resolution $3,700\times3,200$), six raw camera images (each $1,600 \times 1,200$), alignments in FLAME topology & full head including face, neck, ears & 12 individuals $\times$ 12 extreme expression sequences &  3dMD \\ \hline
VOCASET~\cite{VOCA2019} & triangle mesh (80k-140k vertices), texture images (avg resolution $3,700\times3,200$), six raw camera images (each $1,600 \times 1,200$), alignments in FLAME topology  & full head including face, neck, ears, speech & 12 individuals $\times$ 40 dynamic sequences (speech-4D) & 3dMD \\

\bottomrule
\end{tabular}
}
\caption{Overview of publicly available 3D shape and/or appearance scans of human faces.}
\label{table:datasets}
\end{table*}

\subsection{Open challenges}\label{subsec:capchallenges}

The field of face capture is far ahead of face modeling in general and 3DMMs in particular. There is a large gap between the quality of data that can be captured and the data actually used to build 3DMMs. There is a further gap between the quality of this already-deficient data and what a 3DMM is able to synthesize (see Section \ref{sec:modelling}). Hence, from the perspective of 3DMMs, the open challenges in capture do not generally relate to improving the acquisition quality, but to the lack of publicly available data. While there is a decent number of datasets publicly available (see Section \ref{sec-datasets}), most of these contain only moderate quality shape data and no appearance information, with the exception of \cite{stratou2011effect}, which consists of 23 identities only. We believe that the lack of high-quality datasets is due to a variety of reasons. On the one hand, high-quality acquisition devices that can capture both shape and appearance are not readily available. Most of them are custom-built, cannot easily be purchased or licensed, and require expert knowledge for operation. On the other hand, acquiring and processing data may be a time and resource-intense effort, since many systems in the research community were not conceived for scalable deployment but for experimental use; slow capture methods are not applicable to young or elderly people, expensive setups are challenging to replicate on a global scale to capture whole populations, and methods requiring very bright illumination makes it unpleasant to be captured with eyes open. Furthermore, most high-quality systems, in particular ones that also measure appearance, generally require controlled lab conditions which makes it difficult to capture large numbers of the general public. Advances in face capture may alleviate some of these issues. 

Additionally, there are many important broader questions related to data acquisition that remain unanswered. How many faces do we really need to capture in order to build a representative (universal) model? How can we ensure we capture natural expressions? Most people are not trained to perform specific expressions (i.e. FACS\footnote{https://en.wikipedia.org/wiki/Facial\_Action\_Coding\_System}), and will have difficulties performing naturally when put in a capture setup, leading to a biased dataset. How should we deal with bias in general and what is the right sampling strategy with respect to age, gender, ethnicity and so on? Are the capture methods themselves biased? For example, capturing faces with very dark skin is challenging for both photometric and geometric methods. Should we accept that we cannot hope to capture sufficiently broad data and therefore rely on synthesizing additional data or using captured data to build a bootstrap model that is refined on large 2D datasets? These approaches are discussed in Section \ref{sec:deeplearning}.


Finally, there are some philosophical and ethical issues to consider. The human face is unique and highly personal. Once a face has been captured in high detail, it is possible to synthesize new images that are almost indistinguishable from photos. If captured datasets are made publicly available, it is very difficult to control the distribution and use of such data. Obtaining proper informed consent is, therefore, both legally and ethically important but perhaps even this does not go far enough, particularly when consent for minors is given by parents. These issues are beyond the expertise of computer graphics and vision researchers and perhaps suggest a need for discussion and debate with other disciplines.



\renewcommand\vec{\mathbf}
\newcommand{\surfaceS}{\mathcal{S}}

\section{Modeling}\label{sec:modelling}

This section outlines how to compute a 3DMM by modeling the variations of digitized 3D human faces. In particular, the following three types of variations are commonly considered. First, geometric variations across different identities are captured in a \emph{shape model}, as outlined in Section~\ref{subsec:shape_models}. Commonly used models include global models, which represent variations of the entire face surface, and local models, which represent variations of facial parts. Second, geometric variations across different facial expressions are captured in an \emph{expression model}, as outlined in Section~\ref{subsec:expression_models}. Commonly used models can be mainly classified into additive and multiplicative models. More recently, nonlinear expression models are starting to be explored. Third, variation in appearance and illumination are captured in a separate \emph{appearance model} as outlined in Section~\ref{subsec:appearance_illumination_models}. 

It is interesting to note that the landmark paper on 3DMMs published 20 years ago~\cite{BlanzVetter1999} proposed first models for all three types of variation that are still commonly used today.

To compute shape, expression, or appearance models, statistics are performed over a database of face data, where traditionally 3D scans of faces were used, and more recently some approaches also learn face models directly from 2D images, as outlined in Section~\ref{sec:3Dfrom2D}.
This computation of statistics requires correspondence information, that is, anatomically corresponding parts of the faces need to be compared, and hence known either explicitly or implicitly. An overview of how correspondence information is computed for faces is given in Section~\ref{subsec:correspondence}. The most commonly used approach is to compute correspondence information explicitly before computing the 3DMM. Some recent methods compute correspondence information at the same time while the 3DMM is built.

3DMMs are generative models and the ability to synthesize novel faces is a key feature and briefly discussed in Section~\ref{subsec:synth_models}. Finally, this section provides a list of available models and discusses open challenges on 3D face modeling in Sections~\ref{subsec:public_models} and~\ref{subsec:challenges_models}, respectively.

\subsection{Shape models}
\label{subsec:shape_models}
This section considers modeling geometric variation across different subjects computed using classical modelling approaches that use 3D data. To use a set of 3D scans as training data, we require a distance measure between any pair of scans, and computing a distance between raw scans consisting of different numbers of unstructured vertices is a complex problem. Most commonly, the community proceeds by first pre-processing the dataset by deforming a template mesh to all scans, which establishes anatomic correspondences between the points of the scans (see Section~\ref{subsec:correspondence}). We denote the surface of such a pre-processed mesh by $\surfaceS$ in the following. The $i$-th vertex of $\surfaceS$ is denoted by $\mathbf{v}_i\in\R^3$, and its associated vector $\vec{c} \in \mathbb{R}^{3n}$ contains the coordinates of $\mathbf{v}_i$ in a fixed order. All meshes share a common triangulation. We denote the $i$th triangle by  $\mathbf{t}_i=(t^1_i,t^2_i,t^3_i)\in\{1,\dots,n\}^3$, where $t^1_i,t^2_i,t^3_i$ provide indices to the associated vertices $\mathbf{v}_{t^1_i},\mathbf{v}_{t^2_i},\mathbf{v}_{t^3_i}$,  and we denote the complete triangulation by $\mathcal{T}=(\mathbf{t}_1,\dots,\mathbf{t}_m)$. Distances between shapes $\surfaceS_1$ and $\surfaceS_2$ are computed as difference between $\vec{c}_1$ and $\vec{c}_2$ after rigidly aligning $\surfaceS_1$ and $\surfaceS_2$ in $\mathbb{R}^3$.

3DMMs most often follow~\citet{Dryden2002} for their definition of \emph{shape} $\surfaceS$ as containing the geometric information remaining after having removed differences caused by translation, rotation, and sometimes uniform scaling. While scaling is typically not removed for human faces, this is often done in geometric morphometrics (e.g.,~\cite[Section 2]{Dryden2002}).

A \emph{shape space} is traditionally defined as the set of all configurations of $n$ vertices in $\mathbb{R}^3$ with fixed connectivity. Since we are interested in modeling human faces only in the context of 3DMMs, in the following, the term shape space refers to a $d$-dimensional parameter space (with $d \ll n$) that represents \emph{plausible} 3D human faces. In this way, each 3D face has an associated parameter vector $\vec{w} \in \mathbb{R}^d$. 

In 3DMMS, statistical shape analysis is used as generative model, i.e. the \emph{shape space} has an associated probability distribution called prior that is defined by a density function $f(\vec{w})$ and that measures the likelihood that a realistic 3D face would be represented by a particular vector $\vec{w}$ in \emph{shape space}. With a slight abuse of notation, we interpret $\vec{c}$ as a generator function in the following as

\begin{equation}
    \vec{c}:\mathbb{R}^d\rightarrow\mathbb{R}^{3n}
    \label{eq:generator}
\end{equation}
that maps the low-dimensional parameter vector $\vec{w}$ to the vector of all vertex coordinates $\vec{c}(\vec{w}) \in \mathbb{R}^{3n}$. We again use $\vec{v}_i(\vec{w}) \in \mathbb{R}^{3}$ to refer to the $i$th vertex of the mesh given by $\vec{w}$. While the resolution (number of vertices) of the model is usually fixed, a progressive mesh representation based on edge collapse simplification of the generator function has been considered \cite{PatelSmith2011}.

This part considers the case where all faces in the training data have a similar (typically neutral) expression; generator functions that additionally model varying expressions are discussed in Section~\ref{subsec:expression_models}. As in~\cite{Brunton2014Review}, our discussions distinguishes \emph{global models} that model the entire face or head area from \emph{local models} that perform statistics over localized areas.

\subsubsection{Global models}

Let $\{\surfaceS_i\}_i$ 
denote the training shapes and $\{\vec{c}_i\}_i$  their associated coordinate vectors. The seminal work on 3DMMs~\cite{BlanzVetter1999} proposed a global shape model that uses principal component analysis (PCA) to compute the linear generator function as 
\begin{equation}
     \vec{c}(\vec{w}) = \bar{\vec{c}} + \mathbf{E} \vec{w} \,,
    \label{eq:pca}
\end{equation}
where $\bar{\vec{c}}$ is the mean computed over the training data, $\mathbf{E} \in \mathbb{R}^{3n {\times} d}$ is a matrix that contains the $d$ most dominant eigenvectors  of the covariance matrix computed over the shape differences $\{ \vec{c}_i-\bar{\vec{c}}_i \}$, and $\vec{w}$ is the low-dimensional shape parameter vector. One hypothesis of this model is that training faces can be linearly interpolated to generate new 3D faces. Another hypothesis is that the 3D faces in the reduced parameter space $\mathbb{R}^d$ follow a multivariate normal distribution, which can be directly deduced from the eigenvalues corresponding to $\mathbf{E}$. This implies that the density function $f(\vec{w})$ evaluating the likelihood of the parametric representation $\vec{w}$ in shape space is simply the Mahalanobis distance of $\vec{w}$ to the origin.

The 3DMM was originally computed over 200 subjects and has proven to be useful in a variety of applications thanks to its power to generate plausible shapes, and its simple underlying model. A recent study rebuilds such a model from a very large dataset containing 9,663 3D scans and revisits best practices~\cite{LSFM2016}, demonstrating that the originally proposed generator function for shape remains highly relevant in the research community.

One observation by~\citet{BlanzVetter1999} is that moving the representation vector $\vec{w}$ away from the mean face increases their distinctiveness, eventually leading to caricatures of the identity. In order to model distinctive facial identities,~\citet{patel2016manifold} propose an alternative density function $f(\vec{w})$ based on the following observation. Consider the squared Mahalanobis distances from the mean for a set of $d$-dimensional vectors that follow a multivariate Gaussian distribution. These distances form a $\chi^2_d$-distribution, which has expected value $d$. Hence, to preserve the shape distinctiveness related to identity, Patel and Smith restrict the representation $\vec{w}$ to have Mahalanobis distance $\sqrt{d}$ from the mean. \citet{lewis2014probable} propose a similar argument showing that, even if faces are truly Gaussian distributed (which has been shown for the Basel data by a Kolmogorov Smirnoff test for shape and per-vertex color, where the marginal distribution for the shape is close to a Gaussian~\cite{egger2016copulab}), methods that make the assumption that typical faces lie near the mean are not valid.

Recently, \citet{luthi2018gaussian} proposed a nonlinear shape space that models deformations from the mean as Gaussian processes.


\subsubsection{Local models}

Using a global generator function in Equation~\eqref{eq:generator} is known to lead to representations that do not model fine-scale geometric details. To improve the modeling of important localized areas, such as the eye or nose regions, \citet{BlanzVetter1999} initially experimented manually segmenting the face into regions and learning separate PCA models per region. Their results demonstrate that this localized modeling allows for reconstructions of higher fidelity. This idea has been extended since with representations that achieve much higher accuracy than the global PCA model, and this comes in general at the cost of a less compact representation $\vec{w}$. 

First local models segmented the face manually~\cite{Basso2007,Kakadiaris2007,ter20083d}. \citet{Smet2010} and \citet{Tena2011} propose automatic ways of segmenting the faces into areas based on information learned over the displacements of corresponding vertices in the training set. ~\citet{Brunton2011} propose a model that combines shape variations that are localized in different areas with a multi-resolution framework that uses a wavelet decomposition of the 3D face models. Fine-scale geometric detail can alternatively be modeled using hierarchical pyramids that consider differences between a smooth face and increasingly high-resolution geometry representing e.g., wrinkles~\cite{Golovinskiy2006}.

It is also possible to perform localized analysis using different statistical approaches than PCA. \citet{Neumann2013} propose the use of sparse PCA combined with a group sparsity constraint to identify localized deformation components over the training data. \citet{Ferrari2015} follow a related idea and learn a dictionary of deformation components oversampled regions for the application of face recognition. \citet{Wu2016} combine a local deformation subspace model with an anatomical bone structure that acts as a regularizer of the deformation. The local deformation subspace is computed over overlapping localized patches, and the statistical model explicitly factors the rigid and non-rigid deformations applied to each patch.

\subsection{Expression models}
\label{subsec:expression_models}

As simple linear models similar to the ones described can be used to model expression variation for one subject, this section considers models that capture variations of both identity and expression. Unlike simple linear models learned over a dataset of varying identities and expressions (e.g.,~\cite{booth20173d}), our focus is on models that explicitly decouple the influence of identity and expression by modeling them in separate coefficients. We classify these methods into additive, multiplicative, and nonlinear models, depending on how the two sets of coefficients are combined. 


\subsubsection{Additive models}

Given two shapes of the same subject, one with expression $\vec{c}_{\text{exp}}$ and one neutral shape $\vec{c}_{\text{ne}}$, \citet{BlanzVetter1999} transferred expressions between subjects by adding the expression offsets $\boldsymbol{\Delta}_c := \vec{c}_{\text{exp}} - \vec{c}_{\text{ne}}$ to the neutral shape of another subject. 

Several other methods then built on this idea, and model expression variations as an additive offset to an identity model with a neutral expression. 
Formally, additive models are given by
\begin{equation}
    \vec{c}(\vec{w}^s, \vec{w}^e) = \bar{\vec{c}} + \mathbf{E}^s \vec{w}^s  + \mathbf{E}^e \vec{w}^e,
    \label{eq:additive_exp}
\end{equation}
where $\bar{\vec{c}}$ is a mean, $\mathbf{E}^s$ and $\mathbf{E}^e$ are the matrices of basis vectors of the shape and expression space, and $\vec{w}^s$ and $\vec{w}^e$ are the shape and expression coefficients. Note that the basis vectors of the expression space can be interpreted as a data-driven blendshape model, where the basis vectors are orthogonal and do not carry interpretable semantic meaning in general~\cite{Lewis2014}.

Starting with \citet{blanz2003reanimating}, several methods propose to learn two PCA models, one over shape and one over expression to derive $\mathbf{E}^s$ and $\mathbf{E}^e$, and to compute $\bar{\vec{c}}$ as the mean over training data, either in neutral expression, or as sum of two means (one over shape and one over expression). \citet{blanz2003reanimating} learned the expression space from a single subject captured in multiple expressions. \citet{amberg2008expression} extended this work to include expression data from multiple subjects. This leads to a statistical expression model which does not enable control over specific facial expressions. It is therefore feasible for analysis-by-synthesis tasks but limited for controlling or synthesizing specific interpretable expression variation. \citet{Thies15} use blendshapes as the basis vectors of the expression space. These expression blendshapes are not orthogonal and hence information of different blendshapes are potentially redundant. 

\subsubsection{Multiplicative models}


Another body of work model shape and expression variations in a multiplicative manner. \citet{Li2010} propose a method to adapt a pre-defined blendshape model to a specific subject given a small number of static face scans in different expressions, which provides a personalized facial rig. \citet{Bouaziz13} combine a morphable shape model $\vec{c}(\vec{w^s})$ (Eq.~\ref{eq:pca}) with a set of $d_e$ linear expression transfer operators $\textbf{T}_j: \mathbb{R}^{3n}\rightarrow\mathbb{R}^{3n}$ that transform the neutral shape to generate personalized blendshapes. Formally, this model is defined as
\begin{equation}
 \vec{c}(\vec{w}^s, \vec{w}^e) = \sum_{j=1}^{d_e} w_j^e \textbf{T}_j \left(\vec{c}(\vec{w}^s) + \boldsymbol{\delta}^s \right) + \boldsymbol{\delta}_j^e,
 \label{eq:multiplicative_blenshape_model}
\end{equation}
where $\boldsymbol{\delta}^s$ and $\boldsymbol{\delta}_j^e$ are corrective vectors to adapt the blendshapes to the tracked subject, and $w_j^e$ is the $j$-th coefficient of $\vec{w}^e$.

A commonly used multiplicative model is the multilinear model that extends the idea of PCA of performing a singular value decomposition to tensor data by performing a higher-order tensor decomposition (HOSVD) of 3D face data stacked into a training tensor. In particular, given a training set of different identities all captured in the same set of expressions, the vertex coordinates are stacked into a data tensor on which HOSVD is performed. This allows to model correlations of shape changes caused by identities and expressions. This model was first applied to 3D face modeling by \citet{Vlasic2005}, and can be defined as
\begin{equation}
 \vec{c}(\vec{w}_2, \vec{w}_3) = \mathcal{M} \times_2 \vec{w}^s \times_3 \vec{w}^e,
\end{equation}
where $\mathcal{M} \in \mathbb{R}^{3n \times d_s \times d_e}$ denotes the multilinear model tensor, and $\times_i$ denotes the tensor mode-product. Thanks to its expressiveness and simplicity, this model is being used extensively for various applications~\cite{Mpiperis2008,Dale11,Yang2012,BolkartWuhrer2015,fried2016perspective}. To allow modeling localized variations, the multilinear model has been applied to wavelet coefficients at different levels of detail~\cite{Brunton2014}.

Computing a multilinear model with HOSVD requires a complete tensor of data, where each identity needs to be present in all expressions, and the data need to be in semantic correspondence specified by expression labels. This severely limits the kind of data that can be used for training. Recently, a number of methods have been proposed to address this limitation using an optimization approach~\cite{BolkartWuhrer2016}, a custom tensor decomposition method~\cite{Wang_2017}, and an autoencoder structure~\cite{Abrevaya18}, respectively.

\subsubsection{Nonlinear models}

Facial shape and expression are mostly modeled with a linear subspace, often assuming a Gaussian prior distribution.
Few methods exist to model facial variations with nonlinear transformations.
\citet{FLAME2017} introduce FLAME, an articulated expressive head model that provides nonlinear control over facial expressions by combining jaw articulation with linear expression blenshapes.
\citet{ichim2017} use a muscle activation model driven by physical simulation.
\citet{koppen2018gaussian} instead of a single Gaussian distribution use a Gaussian mixture model to represent facial shape and texture.
In another line of works, \citet{shin2014extraction} capture facial wrinkles in multi-scale maps and nonlinearly transfer them to other faces to enhance realism.

Recently, several deep learning-based models were published that fall into this group of nonlinear models~\cite{Bagautdinov2018, CoMA2018, Lombardi2018deep, tewari2018self, t19fml, Tran2018}. 
Section~\ref{sec:deeplearning} covers these models in more detail.

\subsection{Appearance models}
\label{subsec:appearance_illumination_models}

This section describes approaches for modelling the facial appearance, where we distinguish between \emph{linear} and \emph{nonlinear} models.
The appearance of a face is influenced by its albedo and illumination. However, most 3DMMs do not completely separate these factors, so that oftentimes the illumination is baked into the albedo.
Hence, in the following, we call the problem of statistically capturing this information \emph{appearance modeling}.
The most common way to build an appearance model is by performing statistics on appearance information of the training shapes, where the appearance information is usually either represented in terms of per-vertex values or as a texture in \emph{uv}-space. 

\subsubsection{Linear per-vertex models}
Usually, color information is modeled as a low-dimensional subspace that explains the color variations.
This 
leads to an analogous model to the linear shape model:
\begin{equation}
\mathbf{d}(\mathbf{w}^t)=\bar{\mathbf{d}}+\mathbf{E}^t\mathbf{w}^t,\label{eqn:lineartexmodel}
\end{equation}
where $\bar{\mathbf{d}}$ and $\mathbf{E}^t$ shares the same number of rows as $\bar{\mathbf{c}}$ and $\mathbf{E}$ and $\mathbf{w}^t$ is the low-dimensional texture parameter vector. 

\citet{booth20173d} and \citet{booth20183d} use a convex matrix factorization formulation for learning a per-vertex appearance model from images based on back-projection, where it is assumed that the 3D geometry of the face in the image is known. Their appearance model is not built using the color images directly but rather features computed from the images, for example, SIFT. This brings advantages that the features may be somewhat invariant to illumination changes and also that they depend on a local neighborhood which may widen the basin of convergence. In a similar vein, \citet{wang2009face} construct a linear model of spherical harmonic bases (see Section \ref{sec:synthesis}). This jointly models texture (more precisely diffuse albedo) and fine-scale shape (surface normal orientation) such that appearance under any illumination can be synthesized as a linear function of the basis.
%

\subsubsection{Linear texture-space models}
A downside of per-vertex models is that they require compatible resolutions between the shape and appearance representation. This is rather uncommon in computer graphics, where usually a low(er) resolution geometry model (oftentimes including normals) is used in conjunction with a high(er) resolution 2D texture map. Working with a 2D texture also has other advantages, such as the possibility of using image processing techniques to modify the texture maps. With that, such a representation is also amenable for being processed by convolutional neural networks (CNNs), as will be addressed in the next section.

We now turn our attention towards works that build linear appearance models in texture space. The original work by \citet{BlanzVetter1999} used a texture-based representation by representing the face in a cylindrical way. Later, texture-based representations were used to add textural details like wrinkles \cite{paysan2010thesis}, or to segment skin and detect moles \cite{pierrard2008skin}.
\
\citet{D3DFACS2011} model appearance variation in \emph{uv}-space based on sequences of facial images recorded from different views. The images of the dynamic sequences are aligned based on a non-rigid registration so that the color variation can be modeled using a linear subspace model based on PCA. \citet{LYHM2017} also use a \emph{uv}-space appearance representation that is defined for the entire head. 
\citet{Huber2016} use a per-vertex appearance variation model based on PCA, but in addition, also define a common \emph{uv}-mapping so that the model can be textured based on given facial images. \citet{moschoglou2018} formulate a robust matrix factorization problem in order to learn attributed facial \emph{uv}-maps from a collection of training textures. A study on the effect of different \emph{uv}-space embeddings of the texture was presented by \citet{booth2014}.



\subsubsection{Nonlinear models} 
Traditionally, the facial appearance is modeled as a linear subspace, where oftentimes a Gaussian distribution is assumed.
However, as empirically shown by \citet{egger2016copula}, the Gaussian assumption is not very accurate and may lead to a sub-optimal facial appearance model. Hence, the authors proposed to replace a PCA-based appearance model with a \emph{Copula Component Analysis} model~\cite{han2012semiparametric}. Subsequently, this idea was extended to jointly model facial shape, texture, and attributes~\cite{egger2016copulab}. Recent work learned a joint shape and texture model using neural networks with an adverserial loss \cite{gecer2019synthesizing}.  
\citet{alotaibi2017biophysical} use the observation that skin color forms a nonlinear manifold in RGB space, approximately spanned by the colors of the pigments melanin and hemoglobin. They inverse render maps of these parameters and then construct a linear statistical model in the parameter space. The resulting biophysical 3DMM is guaranteed to produce plausible skin colors.
In addition to global facial appearance models, there are also approaches that consider models of local skin variations.
For example, \citet{dessein2015} use a texture model based on small overlapping patches that are extracted from a face database, and \citet{schneider2018} have presented a stochastic model that is able to synthesize freckles.

More recently, a range of appearance modeling approaches based on deep learning have been proposed, where many of these methods are also built within an analysis-by-synthesis framework. These aspects will be discussed in-depth in Secs.~\ref{sec:AbyS} and~\ref{sec:deeplearning}.

\subsection{Joint shape and appearance models}

\citet{BlanzVetter1999} originally proposed building separate, independent models for shape and texture. Interestingly, in 2D the Active Appearance Model \cite{Cootesetal98} was originally proposed with a combined shape and appearance model. The advantage of such a joint model is that correlations between shape and texture can be learned and exploited as a constraint during fitting with fewer parameters. On the other hand, separate models are more flexible and, since shape and texture parameters can be adjusted independently, sequential algorithms can fit the two models independently. However, 3DMMs that jointly model shape and texture have subsequently been considered. \citet{schumacher2015exploration} use canonical correlation analysis to study shape/texture correlations and also correlations between face parts. \citet{egger2016copula} use copula component analysis that can deal with the different scales of shape and texture data. \citet{Zhou2019} propose a deep convolutional colored mesh autoencoder that learns a joint nonlinear model of shape and texture.

\subsection{Correspondence}
\label{subsec:correspondence}


The previously discussed models typically require the data with point-to-point correspondence between all shapes. 
We refer to the process of establishing such a dense correspondence between scans as registration in the following.

Many methods exist to establish point-to-point correspondence for general classes of objects (e.g.,~\cite{Tam2013, VanKaick2011}), yet the space of face deformations is strongly constrained.
Most commonly used face registration methods follow the principle of deforming a template mesh to each scan in the dataset. 
This registration process typically starts with a rough alignment (often using sparse correspondences) and leads to dense correspondences in the end.

While several image-based methods can also be seen as jointly learning correspondence (between images) and building a statistical model (e.g.,~\cite{Tran2018, t19fml}), we cover such deep learning-based methods in Section~\ref{sec:deeplearning} in more details.

\subsubsection{Sparse correspondence computation}


Several methods exist to establish a sparse correspondence for a dataset of 3D scans by predicting landmarks, i.e. a common set of salient points, for each scan.
This sparse correspondence then typically serves as automatic initialization for dense correspondence methods.

Most of the methods use some local descriptors, or combination of local descriptors and connectivity information between descriptors, to predict salient points.
While landmark localization in images is widely researched (e.g.,~\citet{BulatTzimiropoulos2017}), our focus is on methods that establish sparse correspondence between 3D scans.

Existing methods use combinations of different geometric descriptors.
\citet{Passalis2011} use shape index and spin image features, \citet{Berretti2011} use curvature and scale-invariant feature transform (SIFT) features, and \citet{Creusot2013} consider combinations of local features such as Gaussian curvature, mean curvature, and a volumetric descriptor, and learn the statistical distribution of these descriptors for each landmark.

Further, existing methods use geometric relations or relations between landmarks along with geometric feature descriptors.
\citet{Guo2013} project a scan into an image and predict landmarks with a 2D PCA model and geometric relations with additional texture constraints.
\citet{Salazar2014}, similarly to \citet{Creusot2013}, learn the statistical distribution of local surface descriptors with additional Markov network to additionally consider connections between landmarks.
\citet{BolkartWuhrer2015} extend this further to sequences by additionally considering temporal edges within the Markov network.

\subsubsection{Dense correspondence computation}

%


Methods that deform a template to establish correspondence mostly differ in the parameterization of the deformation. 
We group existing methods according to the type of scan data they register. We distinguish here between static methods, i.e. methods that aim at registering static 3D scan, and dynamic methods that register 3D motion sequences. 
\citet{BlanzVetter1999} use a bootstrapping approach to iteratively fit a 3DMM to a scan, refine the correspondence between the model fit and the scan with a flow field, and refine the model.
\citet{blanz2003reanimating} later extend this approach to expressive scans. 
\citet{amberg2008expression} register expressive scans with a non-rigid ICP. 
\citet{Hutton2001} establish a thin-plate spline (TPS) mapping to warp each scan to a reference and establish correspondence using nearest neighbor search. 

\citet{Passalis2011} register scans by deforming an annotated face model (AFM)~\cite{Kakadiaris2005}, i.e. an average 3D face template that is segmented into different annotated areas, by solving a second-order different equation.
\citet{Mpiperis2008} initially fit a subdivision surface to a scan, where the deformation of the base mesh (i.e. the mesh of the lowest subdivision level) is guided by a sparse landmark correspondence.
After registering a training set, they parametrize the deformation of the base mesh with a PCA model over the training data.
\citet{Salazar2014} use a generic expression blendshape model to fit the expression of the scan, followed by a non-rigid ICP to closely fit the surface of the scan.
\cite{bfm17} establish dense correspondence with a Gaussian process deformation model with the spatially varying kernel.

Several methods exist to sequentially register motion sequences. 
\citet{Weise09} use an identity PCA model to register a neutral scan, and then track motion sequences by optimizing sparse and dense optical flow between consecutive frames.
\citet{Fang2012} and \citet{FLAME2017} initialize the optimization by the registration of the previous frame to exploit temporal information. 
\citet{Fang2012} use an AFM, \citet{FLAME2017} a non-rigid ICP regularized by FLAME. 
\citet{Fernandez2018} uses a spatiotemporal method to register entire motion sequences by iteratively refining the registration of entire sequences by explicitly encoding temporal information with a Discrete Cosine Transform (DCT).

Further, non-template fitting based registration methods exist. 
\citet{Sun2010} use a conformal mapping to parameterize two meshes and establish dense correspondence between the resulting planar meshes by extrapolating sparse landmark correspondences.
\citet{Ferrari2015} segment the face scans into non-overlapping parts divided by geodesic curves between selected landmark pairs, and consistently re-sample each part.

\subsubsection{Jointly solving for correspondence and statistical model}



\citet{Li13} and \citet{Bouaziz13} jointly update person-specific blendshape models and register motion sequences in a real-time facial animation framework.
During tracking, \citet{Li13} use an adaptive PCA model that combines the person-specific blendshapes with additional corrective basis vectors that are successively updated, and \citet{Bouaziz13} optimizes for corrective deformation fields (Equation~\ref{eq:multiplicative_blenshape_model}).

\citet{BolkartWuhrer2015ICCV} and \citet{Zhang2016} instead optimize correspondence for a dataset of different subjects in multiple expression in a groupwise fashion.
\citet{BolkartWuhrer2015ICCV} jointly update the point correspondence within the mesh surface by minimizing an objective function that measures the compactness of a multilinear face model.
\citet{Zhang2016} optimize functional maps across the entire dataset.

\subsection{Synthesis of novel model instances}\label{subsec:synthesis}
\label{subsec:synth_models}

3DMMs can be used to synthesize new 3D faces that are different from any of the observed training data, yet realistic. This can be achieved by altering the coefficients in parameter space (i.e. shape space, expression space or appearance. Common operations in parameter space include interpolating or extrapolating between the coefficients of training samples. Furthermore, any of the generative models presented in this section can be used to directly synthesize new 3D faces by drawing random samples in parameter space according to the prior distribution. Depending on the model, this sampling allows to synthesize or alter identity, expression, or appearance of a static 3D face. Synthesis works are heavily used for entertainment purposes, and these works are discussed in Section~\ref{sec:app:entertainment}. 

Synthesis of static 3D faces notably includes the generation of face caricatures by moving the identity coefficient linearly away from the mean~\cite{BlanzVetter1999} which is mainly explored to study the human face processing system as discussed in Section~\ref{sec:app:humanFaceProcessing}. 

With 3DMMs that encode and decouple identity and expression information, it is easy to synthesize dynamic sequences by fixing the identity coefficients while modifying the expression coefficients. Some works aim to synthesize coherent dynamic 3D face videos of a fixed identity with the help of 3DMMs. These include works that synthesize 4D videos from a static 3D mesh paired with semantic label information~\cite{BolkartWuhrer2015}, and from a static 3D mesh and audio information~\cite{VOCA2019}.

\subsection{Publicly available models}
\label{subsec:public_models}

\begin{figure*}
    \centering
    \includegraphics[width=0.9\linewidth]{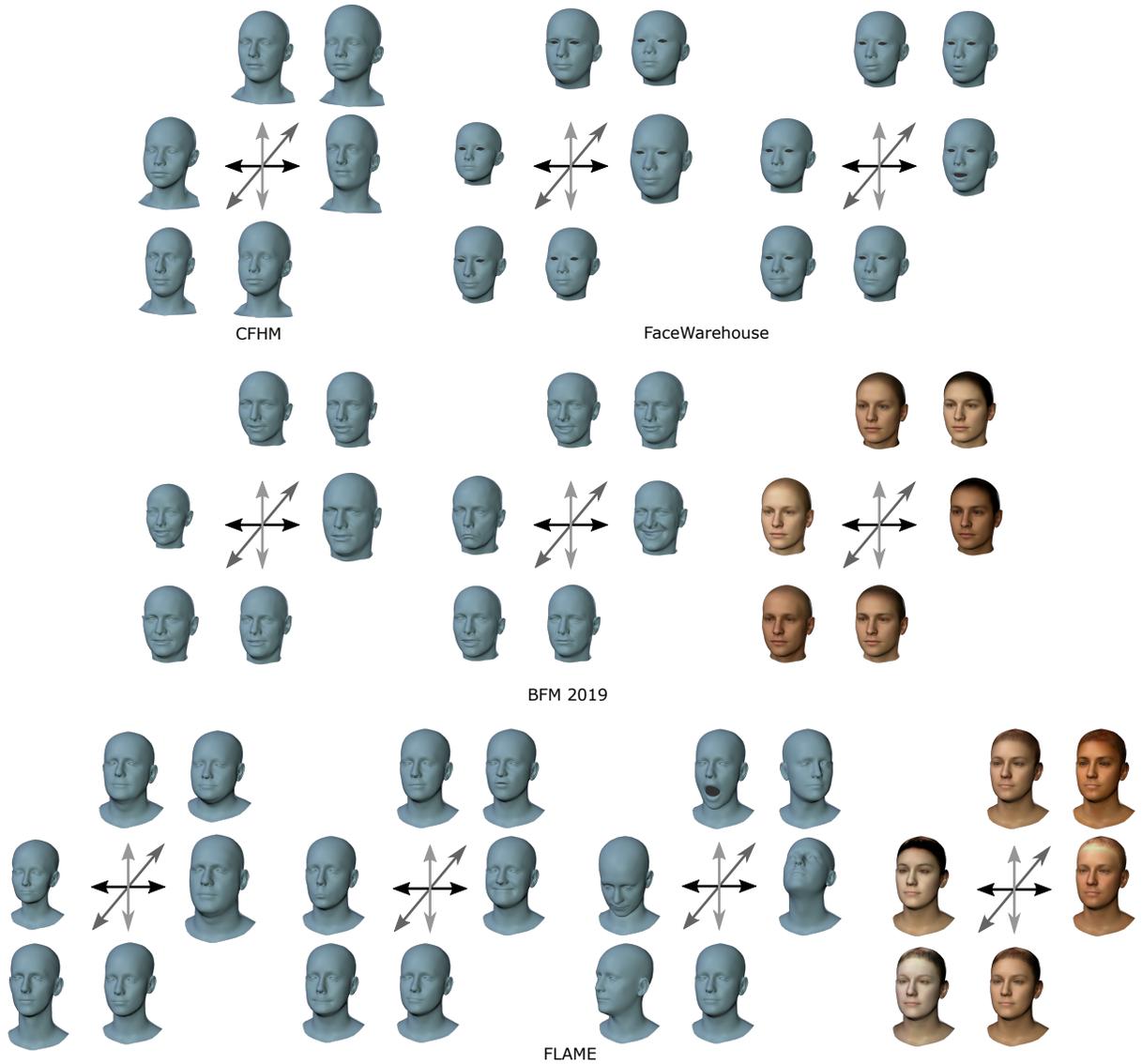}
    \caption{Model variations of existing face models. Top left: CFHM~\cite{Ploumpis_2019_CVPR} shape variations. Top right: FaceWarehouse~\cite{Cao2014_FaceWarehouse} shape and expression variations (while the original model is not available to the best of our knowledge, the visualized multilinear face model is trained from the published FaceWarehouse dataset). Middle:  BFM 2019~\cite{bfm17} shape, expression, and appearance variations. Bottom: FLAME~\cite{FLAME2017} shape, expression, pose, and appearance variation. For shape, expression, and appearance variations, three principal components are visualized at $\pm 3$ standard deviations. The FLAME pose variations are visualized at $\pm \pi/6$ (components three and four) and at $0, \pi/8$ (component six).}
    \label{fig:models}
\end{figure*}

In Table~\ref{table:models} we list publicly available shape and/or appearance models of human faces. Figure~\ref{fig:models} visualizes geometry or appearance variations of some models. We also refer to the curated list of 3DMM software and data that we collected, share and update \cite{OverviewGithub2019}.

\begin{table*}[h!]
\begin{tabular}{m{0.27\linewidth} m{0.15\linewidth} m{0.1\linewidth} m{0.18\linewidth} m{0.1\linewidth}} 
\toprule
\textbf{model} & \textbf{geometry} & \textbf{appearance} & \textbf{data} & \textbf{comment} \\ \midrule
Basel Face Model (BFM) 2009 $\newline$\cite{paysan2009face} & shape & per-vertex & 200 individuals, each in neutral expression & includes separate models for facial parts\\ \hline
FaceWarehouse \cite{Cao2014_FaceWarehouse} & shape, expression & - & 150 individuals, each with 20 expressions &  \\ \hline
Global and local linear model $\newline$\cite{Brunton2014Review} & shape & - & 100 individuals &  \\ \hline
Multilinear Wavelet model $\newline$\cite{Brunton2014} & shape, expression & - & 99 individuals, 25 expressions 
&  \\ \hline
Multilinear face model $\newline$\cite{BolkartWuhrer2015ICCV} & shape, expression & - &  2500 scans (100 individuals, 25 expressions)
&  \\ \hline
Multilinear face model $\newline$\cite{BolkartWuhrer2016} & shape, expression & - &  2510 scans (205 individuals, up to 23 expressions)&  \\ \hline
Large Scale Facial Model (LSFM)$\newline$\cite{LSFM2016} & shape & - & 9663 individuals &  \\ \hline
Surrey Face Model $\newline$\cite{Huber2016} & shape, expression & per-vertex  & 169 individuals & multi-resolution \\ \hline

Liverpool-York Head Model (LYHM) $\newline$\cite{LYHM2017} & shape & per-vertex& 1212 individuals & full head (no hair, no eyes) \\ \hline
Faces Learned with an Articulated Model and Expressions (FLAME)\newline\cite{FLAME2017} & shape, expression, head pose & texture & 3800 individuals for shape, 8000 for head pose, 21000 frames for expression & female, male, gender neutral model, full head (no hair) \\ \hline
Basel Face Model (BFM) 2017 \newline \cite{bfm17} & shape, expression & per-vertex & 200 individuals for shape and appearance, a total of 160 expression scans & BFM 2019 with full head and multi-resolution \\ \hline
York Ear Model \cite{dai2018data} & shape & - & 20 3D ear scans, augmented with 605 landmark-annotated 2D ear images & ear only \\  \hline
Multilinear autoencoder $\newline$\cite{Abrevaya18} & shape, expression & - & 5000 scans from 195 individuals, 500000 after augmentation &  \\ \hline
Convolutional Mesh Autoencoder (CoMA) \cite{CoMA2018} & shape, expression & - & 12 individuals, 12 extreme expressions, 20466 meshes in total & full head (no hair) \\  \hline
Combined Face \& Head Model (CFHM) \cite{Ploumpis_2019_CVPR} & shape  & - & Merged from LYHM and LSFM models & full head (no hair) \\ \hline
Morphable Face Albedo Model $\newline$ \cite{smith2020albedo} & - & per-vertex diffuse and specular albedo & 73 individuals (50 scanned + 23 3DRFE \cite{stratou2011effect}) & extends BFM2017\\ 
\bottomrule
\end{tabular}
\caption{Overview of publicly available 3D shape and/or appearance models of human faces.}
\label{table:models}
\end{table*}




\subsection{Open challenges}
\label{subsec:challenges_models}

While 3D face modeling has received considerable attention during the past two decades, some challenges remain. First, the statistics of most models are limited to the face and do not include information on eyes, mouth interior or hair. These details are however crucial for many applications, and it is not straightforward to combine a 3DMM with specific models e.g., for hair. Second, the interpretability of the representations would benefit from being improved. PCA is the most commonly used method to perform statistics on 3D faces, and as it is an unsupervised method, the principal components do not coincide with attributes that humans would use to describe a face. Third, methods that incorporate different levels of detail typically come at the cost of a less compact representation, and it is unknown how many parameters are required to accurately represent facial geometry and appearance at varying levels of detail. Fourth, the different models presented in this section have different advantages and drawbacks, making them most suitable for specific applications. It is unknown whether one integrated optimal model for all applications exists. Fifth, all currently available models, even the large scale ones have a very strong racial bias towards white. This can be alleviated in the future by scanning efforts in different parts of the world. Another potential solution to overcome a racial bias can be the generative model itself, as these allow to generate and add synthetic data to biased datasets. Sixth, learning from inhomogeneous data presents another open challenge. There are many available datasets with different resolution, coverage, noise characteristics, biases and so on (see Table \ref{table:datasets}). To make the best use of this data requires methods that can learn models from all data sources simultaneously but this requires explicit ways to deal with data inhomogeneity. Some very recent work begins to look at this problem \cite{liu20193d}. Finally, there are some fundamental open questions related to the statistical modeling of shape. Two face shapes differ by nonlinear shape deformation superposed on top of rigid body motion. Conventionally, this is dealt with by first rigidly aligning, then modeling the residual shape differences but this makes the model dependent on the choice of alignment metric. For faces specifically, estimated skull position has been used for rigid alignment \cite{beeler2014rigid}. Although not applied to faces, a recent method uses a rigid body motion invariant distance measure to learn nonlinear principal components \cite{heeren2018principal}.


\section{Image Formation}
\label{sec:synthesis}


A 3DMM provides a parametric representation of face geometry and appearance. One key usage of such a model is synthesis, which involves two steps. First, generating a new model instance via sampling from the parameter space or manual interaction with model parameters (see Section \ref{subsec:synthesis}). Second, rendering the generated model into a 2D image via a simulation of the image formation process, i.e.~the computer graphics pipeline. The synthesis also forms an important part of 3DMM-based face analysis, either through classical analysis-by-synthesis (see Section \ref{sec:AbyS}) or as a model-based decoder within a deep learning architecture (see Section \ref{sec:deeplearning}). In this section, we focus on modeling the image formation process. This potentially encompasses the whole of the rendering literature, so we restrict our attention to techniques and models that have been applied in the context of 3DMMs. We cover the geometry and photometry of image formation in Sections \ref{sec:synthesis:cameras} and \ref{sec:synthesis:photometric}, the rendering pipelines used for 3DMM fitting in Section \ref{sec:synthesis:rendering} and finally in Section \ref{sec:synthesis:open} we highlight where there are future opportunities for exploiting state-of-the-art rendering techniques to improve 3DMM synthesis.

\subsection{Geometric image formation}\label{sec:synthesis:cameras} 

A camera model describes the \emph{geometry} of image formation, specifically, how positions in the 3D world project to 2D locations in the image plane. A variety of camera models have been used in the 3DMM literature which are described here in order of increasing accuracy with respect to a real camera. We denote the projection of a 3D point $\mathbf{v}=[u, v, w]^{\textrm{T}}$ onto the 2D point $\mathbf{x}=[x, y]^{\textrm{T}}$ by $\mathbf{x}=\textrm{\textbf{project}}[\mathcal{C},\mathbf{v}] \in \R^2$, where $\mathbf{project}$ represents one of the camera projection models below and $\mathcal{C}=(\mathcal{C}_{\text{intrinsic}},\mathcal{C}_{\text{extrinsic}})$ contains the camera parameters. $\mathcal{C}_{\text{extrinsic}}=(\mathbf{R},\mathbf{t})$ describes the pose in terms of a rotation $\mathbf{R}\in SO(3)$ and translation $\mathbf{t}\in\R^3$ that transform from world to camera coordinates. $\mathcal{C}_{\text{intrinsic}}$ is a set of internal parameters specific to each projection model. The task of estimating $\mathcal{C}$ is known as camera calibration or camera resectioning and is usually done from known or estimated 2D-3D correspondences. Estimating $\mathcal{C}_{\text{extrinsic}}$ with known $\mathcal{C}_{\text{intrinsic}}$ is called pose estimation or, in the case of a perspective camera model, the perspective-n-point problem.

\paragraph{Scaled orthographic}
The scaled orthographic projection model comprises an orthographic projection whose sole parameter is a uniform scaling $s\in\R_{>0}$:
\begin{equation}
 \textrm{\textbf{ortho}}[\mathbf{v},\mathbf{R},\mathbf{t},s] = s\mathbf{P}(\mathbf{Rv}+\mathbf{t}) = s\begin{bmatrix}\mathbf{r}_1 & t_1\\\mathbf{r}_2 & t_2\end{bmatrix}\tilde{\mathbf{v}} = \mathbf{C}\tilde{\mathbf{v}},\quad
\mathbf{P}=
\begin{bmatrix}
 1 & 0 & 0 \\
 0 & 1 & 0
\end{bmatrix}
\end{equation}
where $\tilde{\mathbf{v}}=[u, v, w, 1]^{\textrm{T}}$ is the homogeneous representation of $\mathbf{v}$ and $\mathbf{r}_1$, $\mathbf{r}_2$ are the first two rows of $\mathbf{R}$. This model is not physically meaningful but is linear in vertex position, translation and scale and avoids size/distance/perspective ambiguities introduced by more realistic camera models. Since $\mathbf{R}$ must be restricted to $SO(3)$, the projection is nonlinear in any parameterization of $\mathbf{R}$. In the context of 3DMMs, scaled orthographic projection has been used for example by \citet{Blanz:04b,Knothe:06,PatelSmith2009,bas2016fitting}. The scaled orthographic model can be interpreted as an approximation to perspective projection when the distance between the surface and camera is large relative to the depth variation. Concretely, when $\max_w-\min_w \ll \bar{w}$ with $\bar{w}=\text{mean}_w$ the mean distance between the surface and the camera, then the nonlinear division in perspective projection can be approximated by a fixed scale $s=f/\bar{w}$ where $f$ is the focal length of the camera. This gives physical meaning to the scaled orthographic model.

\paragraph{Affine} The affine camera generalizes the orthographic model by allowing arbitrary affine transformations. Specifically, this additionally allows non-uniform scaling and skew transformations which can approximate perspective effects whilst remaining linear. An affine camera can be represented by an arbitrary matrix $\mathbf{C}\in\R^{2\times 4}$ with the projection given simply by $\mathbf{x}=\textrm{\textbf{affine}}[\mathbf{v},\mathbf{C}]=\mathbf{C}\tilde{\mathbf{v}}$. The projection is linear in $\mathbf{C}$ and since its 8 entries are unconstrained, they can be estimated using linear least squares (though note that numerical stability entails first performing a normalization procedure). In the context of 3DMMs, the affine camera has been used for example by \citet{aldrian2013inverse,Huber2016}.

\paragraph{Perspective}
A nonlinear perspective projection is given by the pinhole camera model $\mathbf{x}=\textrm{\textbf{pinhole}}[\mathbf{v},\mathbf{K},\mathbf{R},\mathbf{t}]$. The matrix:
\begin{equation*}
\mathbf{K}=\begin{bmatrix}
f_x & \gamma & c_x  \\
0 & f_y & c_y  \\
0 & 0 & 1 
\end{bmatrix}
\end{equation*}
contains the intrinsic parameters of the camera, namely the focal length in the $x$ and $y$ directions $f_x,f_y\in\R_{>0}$, the skew $\gamma\in\R$ and the principal point $[c_x,c_y]\in\R^2$. Common assumptions are that the pixels are square (in which case a single focal length $f=f_x=f_y$ parameter is used), that the camera sensor is perpendicular to the camera view vector (in which case $\gamma=0$) and that the principal point is in the centre of the image ($c_x=w/2$ and $c_y=h/2$). Note that $f$ is actually a product of two quantities: the physical focal length in world units (e.g.,~mm) and the conversion factor from world units to pixels (i.e.~with units of pixels/mm). The nonlinear perspective projection can be written in linear terms by using homogeneous representations:
\begin{equation}
 \lambda\tilde{\mathbf{x}}=\mathbf{K}
\begin{bmatrix}
\mathbf{R} & \mathbf{t}
\end{bmatrix}
\tilde{\mathbf{v}}=\mathbf{C}\tilde{\mathbf{v}},\label{eqn:linearcammodel}
\end{equation}
where $\tilde{\mathbf{x}}=[x, y, 1]^{\textrm{T}}$, $\lambda$ is an arbitrary scaling factor and $\mathbf{C}\in\R^{3\times 4}$ is known as the camera matrix. The final image coordinate is obtained by the nonlinear homogenization of $\tilde{\mathbf{x}}$. Unlike the linear models, the pinhole model captures the effect of distance on projected shape. This becomes important when a face is close to the camera. At ``selfie'' distance (e.g.,~0.5m), the difference between perspective and orthographic projection of 3D face landmarks is about 6\% of the interocular distance \cite{bas2017does}. For this reason perspective projection is commonly used in the context of 3DMMs, for example, in the original \citet{BlanzVetter1999} paper and more recently in a shape-from-landmarks setting \cite{Cao2013,Cao2014,Saito2016}. Unfortunately, since calibration information is rarely available the increased complexity of this model introduces ambiguities between shape, scale and focal length that have only recently been studied \cite{smith2016perspective,bas2017does}, though the ambiguity has often been hinted at in the literature. For example, the original \citet{BlanzVetter1999} paper relied on a fixed, manually provided subject-camera distance. \citet{booth20183d} state ``we found that it is beneficial to keep the focal length constant in most cases, due to its ambiguity with $t_z$''. \citet{Schoenborn2017} explored the ambiguity of estimated distance from the camera under perspective and observed a very high posterior standard deviation and the distance can not be resolved even by using a strong prior for the face shape. In \citet{tewari17MoFA}, a similar effect is observed, indicated by the learning rate on the z translation (i.e.~subject-camera distance), which is set three orders of magnitude lower than all other parameters. Both approaches in practice fix the face distance to avoid this difficulty.

\subsection{Photometric image formation}\label{sec:synthesis:photometric} 


The appearance of a face is determined by the interaction of light with the material of the face, predominately skin. Hence, the \emph{photometry} of both illumination and reflectance must be modeled in order to simulate the image formation process. 

\paragraph{Reflectance models} The reflection of light from a surface is often described using a Bidirectional Reflectance Distribution Function (BRDF). This describes the directional dependence of local light reflection from an opaque surface. It is represented by a four dimensional function $f_r({\bm \omega}_i,{\bm \omega}_o)$ that gives the ratio of \textit{outgoing} reflected radiance in direction ${\bm \omega}_o$ to \textit{incoming} incident irradiance from direction ${\bm \omega}_i$. 
A BRDF allows us to express irradiance $L_o({\bm \omega}_o) $ in direction ${\bm \omega}_o$ as a function of light reflected from all incident directions:
\begin{equation}
    L_o({\bm \omega}_o) = 
    \int_{\Omega(\mathbf{n})} f_r({\bm \omega}_i,{\bm \omega}_o) L_i({\bm \omega}_i) \cos\theta_i \text{d}{\bm \omega}_i,\label{eqn:outgoingradiance}
\end{equation}
where $L_i({\bm \omega}_i)$ is incident irradiance from direction ${\bm \omega_i}$, $\Omega(\mathbf{n})$ is the hemisphere around the local surface normal $\mathbf{n}$ and $\theta_i$ is the angle between ${\bm \omega}_i$ and $\mathbf{n}$. Note $\cos\theta_i=\mathbf{n}\cdot {\bm \omega}_i$, where $\cdot$ denotes the inner product.
Physically-valid BRDFs must exhibit a number of properties: nonnegativity ($f_r({\bm \omega}_i,{\bm \omega}_o)\geq 0$), Helmholtz reciprocity ($f_r({\bm \omega}_i,{\bm \omega}_o)=f_r({\bm \omega}_o,{\bm \omega}_i)$) and conservation of energy:
\begin{equation}
    \forall {\bm \omega}_i, \int_{\Omega(\mathbf{n})} f_r({\bm \omega}_i,{\bm \omega}_o)\cos\theta_i\text{d}{\bm \omega}_o\leq 1.
\end{equation}
A particularly simple and commonly used physically valid BRDF is the Lambertian model for a perfectly diffuse reflector. The Lambertian model assumes incident light is scattered equally in all directions resulting in a constant BRDF: $f_{\text{Lambert}}({\bm \omega}_i,{\bm \omega}_o)=\rho_d/\pi$. $\rho_d\in[0,1]$ is the diffuse reflectivity or \emph{albedo}, that is usually wavelength dependent and can be thought of as the color of the object. Work predating the original 3DMM used a linear statistical 3D face shape model with the Lambertian reflectance model in a shape-from-shading context \cite{atick1996statistical}. Subsequently, the Lambertian model has been used for 3DMM fitting in the context of the spherical harmonic lighting model \cite{zhang2006face} (see below) where its simplicity yields closed-form expressions. This is now very common, including in the current state-of-the-art, e.g.,~\cite{tran2018learning,tran2019towards}. In general, the Lambertian model is a poor approximation for the complex reflectance properties of facial skin, hair, eyes, etc., and so more sophisticated models have been considered. 

\citet{BlanzVetter1999}  originally used the Phong model which augments the Lambertian term with a constant ambient term and a phenomenological specular model enabling the simulation of glossy reflectance. The Phong model can be described in terms of the following BRDF:
\begin{equation}
    f_{\text{Phong}}({\bm \omega}_i,{\bm \omega}_o)=\frac{\rho_a+\rho_s(\mathbf{r}\cdot{\omega_o})^\eta}{\mathbf{n}\cdot{\bm \omega_i}}+\rho_d
\end{equation}
where $\mathbf{r}$ is the reflection of ${\bm \omega}_i$ about $\mathbf{n}$, $\eta$ is the \emph{shininess} that controls the width of the specular lobe and $\rho_a$, $\rho_s$ are ambient and specular ``albedos''. In the context of 3DMMs, usually only $\rho_d$ is allowed to vary spatially. Note that the Phong BRDF does not satisfy the constraints above for physical validity. In graphics, extremely complex, physically-valid BRDF models have been developed specifically for materials of relevance to face 3DMMs, for example for skin \cite{krishnaswamy2004biophysically} and hair \cite{marschner2003light}. Note that skin is a layered, partially translucent material and so a local BRDF model is inadequate to describe the actual subsurface scattering effects that take place. More complex 8-dimensional bidirectional subsurface scattering reflectance distribution functions (BSSRDF) have been proposed for such materials. However, both these and the more complex BRDFs have proven to be too complex to integrate into 3DMM fitting pipelines and so the majority of work has used Lambertian or non-physical models of moderate complexity.

\paragraph{Lighting} In \eqref{eqn:outgoingradiance}, $L_i({\bm \omega}_i)$ represents the hemispherical incident illumination environment at the surface point. Natural illumination is usually complex, comprising multiple, possibly extended sources as well as secondary illumination reflected from other surfaces. A common assumption is that the illumination environment is distant relative to the size of the object in which case it can be represented by a constant 2D \emph{environment map}, a discrete approximation of $L_i({\bm \omega}_i)$ that is used for every point on the surface. However, the space of possible natural illumination is very high dimensional and rendering with an environment map is computationally expensive, so a number of further simplifications are commonly used.

The simplest illumination model is a point source, in which $L_i({\bm \omega}_i)$ is a delta function characterized by a unit vector in the light source direction, $\mathbf{s}$, and an intensity, $L_i$. Ignoring constants and assuming image intensity is proportional to surface radiance, we can plug in the simple BRDF models above and obtain the following shading models:
\begin{equation}
    I_{\text{Lambert}}=L_i \rho_d \mathbf{n}\cdot\mathbf{s},\quad I_{\text{Phong}}=L_i\left[\rho_a+\rho_d \mathbf{n}\cdot\mathbf{s}+\rho_s (\mathbf{r}\cdot\mathbf{v})^{\eta}\right],
\end{equation}
where $\mathbf{v}$ is a unit vector in the viewer direction. Usually, the light source intensity and albedos would be RGB values. Often $\rho_a=\rho_d$, representing the intrinsic color of the surface. In a 3DMM, this is described by the statistical texture model \eqref{eqn:lineartexmodel}. Note that these simple models are purely local, this means that they neglect self occlusion of the light source, i.e.~cast shadows. These can be added at the cost of computing the occlusion function which is not differentiable. 

A better approximation of complex natural illumination is provided by the spherical harmonic illumination model \cite{ramamoorthi2001efficient}. 
Spherical harmonics provide an orthonormal basis for functions on the sphere, analogous to a Fourier basis in Euclidean space:
\begin{equation}
    I_{\text{SH}}(\mathbf{n}) = \sum_{l=0}^{\infty}\sum_{m=-l}^l l_{l,m} \mathcal{B}_{l,m}(\mathbf{n}),
\end{equation}
where $\mathcal{B}_{l,m}(\mathbf{n})$ are the orthonormal basis functions, and $l_{l,m}$ are coefficients describing reflectance and illumination. The subscript $l$ denotes the degree and $m$ the order of the spherical harmonics. In the Lambertian case, the contribution from the reflectance function is constant and 98\% of the energy of the reflectance function can be captured for any illumination environment using an order 2 ($l=\{0,1,2\}$) approximation. In practice, this means that a good approximation for appearance can be obtained using 9 illumination coefficients per color channel. Combining this model with the linear texture model for albedo $\mathbf{d}(\mathbf{w}^t)$ yields the following:
\begin{equation}
    \mathbf{i}_{\text{SH}} = \mathbf{d}(\mathbf{w}^t) \odot \text{vec}(\mathbf{B}(\mathbf{w}^s)\mathbf{L}),
\end{equation}
where $\odot$ is the Hadamard (element-wise) product. The matrix $\mathbf{B}(\mathbf{w}^s)\in\R^{n\times 9}$ contains the spherical harmonic basis for each vertex which depends on the vertex normal direction and hence the geometry which in turn is determined by the shape parameter vector. The matrix $\mathbf{L}\in\R^{9\times 3}$ contains the lighting coefficients for each color channel. 
\citet{zhang2006face} were the first to fit this model in the context of 3DMMs. \citet{aldrian2013inverse} additionally used the same model for specular reflection, showing that the coarse structure of the illumination environment can be recovered from a face image. They also introduced priors to help resolve lighting/texture ambiguities. \citet{Egger2018IJCV} went further by learning a low dimensional illumination prior from spherical harmonic lighting coefficients estimated from real in-the-wild images.

An alternate model that takes a step towards capturing global illumination effects is the \emph{ambient occlusion} model. Here, it is assumed that $L_i({\bm \omega}_i)$ is constant everywhere, i.e.~that illumination is perfectly diffuse. In this case, shading depends only upon the degree to which the incident hemisphere is occluded. The ambient occlusion, $A_{\mathbf{v}}$, at vertex $\mathbf{v}$ is given by:
\begin{equation}
A_{\mathbf{v}}=\frac{1}{\pi}\int_{\Omega(\mathbf{n})} V(\mathbf{v},{\bm \omega})(\mathbf{n}\cdot{\bm\omega})\text{d}\omega,
\end{equation}
where $V(\mathbf{v},{\bm \omega})$ is the visibility function defined as zero if vertex $\mathbf{v}$ is occluded in direction ${\bm \omega}$, and one otherwise. One can also define the \emph{bent normal} as the average unoccluded direction. Using the bent normal with the spherical harmonic illumination model and scaling the result by the ambient occlusion provides a rough approximation of global illumination effect. Ambient occlusion and bent normal direction depends on the geometry and hence the 3DMM shape parameters. \citet{aldrian2012inverse} proposed to learn a linear model of ambient occlusion and bent normals and included this in their 3DMM synthesis model. \citet{zivanov2013human} similarly construct a joint linear model of spherical harmonic bases and ambient occlusion.

The most complex global model of appearance considered in the context of 3DMMs is the precomputed radiance transfer (PRT) model \cite{sloan2002precomputed}. This uses an efficient representation (such as spherical harmonics) to approximate the local light transport at each vertex, accounting for shadowing and inter-reflection. These are precomputed but can then be used with any incident illumination at render time. \citet{schneider2017efficient} learn a linear model of PRT transfer matrices as a function of the 3DMM shape coefficients and use this in a rendering framework.

We denote by $\mathbf{L}$ the set of illumination parameters for any of the above illumination models.

\paragraph{Color transformation} In a real camera, the actual image irradiance measured by the sensor is usually transformed in a complex way in order to achieve a pleasing visual appearance. Often, this amounts to multiplication by a $3\times 3$ color transformation matrix followed by a nonlinearity. The color transformation matrix can be decomposed into a product of three $3\times 3$ matrices: $\mathbf{T}=\mathbf{T}_{\textrm{xyz2rgb}}\mathbf{T}_{\textrm{raw2xyz}}\mathbf{T}_{\textrm{wb}}$, where $\mathbf{T}_{\textrm{wb}}$ is a diagonal matrix that performs white balancing (compensating for the color of the illumination), $\mathbf{T}_{\textrm{raw2xyz}}$ is specific to each camera and maps from the native color space to the standardized XYZ space and $\mathbf{T}_{\textrm{xyz2rgb}}$ is a fixed matrix that transforms to sRGB space. Unfortunately, introducing such a color transformation into the 3DMM image formation model further exacerbates the lighting/albedo ambiguity by providing an additional explanation for observed color. Finally, a nonlinearity is applied, which can be approximated by $\mathbf{i}_{\textrm{sRGB}} = \mathbf{i}_{\textrm{linRGB}}^{1/\gamma}$, where usually $\gamma=2.2$. This nonlinear transformation is important because it means the, often linear, reflectance and illumination models described above cannot explain the final image appearance.

Despite their importance, camera color transformations and nonlinear gamma are almost always ignored in the context of 3DMMs. There are some notable exceptions. \citet{schneider2017efficient} apply gamma correction to input images to transform back to a linear space. \citet{blanz2003face} estimate a per-channel scale and offset as well as scalar color contrast, allowing them to synthesize grayscale images. The same model was used by \citet{aldrian2013inverse} and \citet{hu2013facial}.

\subsection{Rendering and visibility} \label{sec:synthesis:rendering} 

3DMM fitting algorithms differ in whether they synthesize a discrete image in image space (i.e.~one color per-pixel) or perform rendering in object space (i.e.~one color per-vertex or per-triangle). The former holds the advantage that it is straightforward to incorporate a texture model built in a high-resolution UV space and also that the output is a regular pixel grid that can be passed to CNN, for example for an adversarial loss \cite{Shamai2019}. Methods that work in object space compute an appearance error by projecting the model vertices into the image and sampling image intensities onto the visible vertices. Visibility can also be computed in either object space or image space, with the latter usually being more efficient.

The original \citet{BlanzVetter1999} paper used object space rendering in which a single color was computed for each triangle center (equivalent to flat shading) with image space z-buffering used for visibility testing. Many subsequent methods also worked in object space but usually with per-vertex colors computed using the reflectance models described above with per-vertex surface normals. This has begun to change recently when more conventional rasterization pipelines have been included in 3DMM synthesis. Rasterization associates with each pixel $(x,y)\in\mathcal{I}$, where $\mathcal{I}=\{1,\dots,w\}\times\{1,\dots,h\}$, a triangle index or a $\texttt{NULL}$ value if the pixel is not covered by a triangle:
\begin{equation}\textbf{raster}_{\mathcal{C},\mathcal{T},\mathbf{w}^s,\mathbf{w}^e}:\mathcal{I}\mapsto\{1,\dots,m,\texttt{NULL}\},
\end{equation}
recalling that $\mathbf{w}^s$, $\mathbf{w}^e$ are the shape and expression parameters respectively and $\mathcal{T}$ the mesh triangulation. Since this is a discrete function it is not smooth and not differentiable. In addition, for each pixel, three weights are calculated that are associated with the vertices of the rasterized triangle: $\mathbf{a}_{\mathcal{C},\mathcal{T},\mathbf{w}^s,\mathbf{w}^e}(x,y)\in\R^3_{\geq 0}$. These weights depend on the projected positions of the vertices
\begin{equation}
    \mathbf{v}_{t^i_{\textbf{raster}_{\mathcal{C},\mathcal{T},\mathbf{w}^s,\mathbf{w}^e}(x,y)}}, i\in\{1,2,3\}.
\end{equation}
Often, these weights are barycentric coordinates of the pixel center within the triangle. These weights are a smooth function of the vertex positions and hence of the shape and camera parameters. Hence, rendering is differentiable up to a change in rasterization, i.e.~so long as the triangle index associated with each pixel does not change. \citet{Tran2018} incorporate such a conventional rasterization pipeline into an in-network differentiable renderer. 

Collecting together all of the parameters relating to the camera, illumination, face geometry and texture, $\Theta=(\mathcal{C},\mathcal{L},\mathbf{w}^s,\mathbf{w}^e,\mathbf{w}^t)$, we can write the rendered appearance in object space of vertex $j$ as $I_{\text{model}}^j(\Theta)$. For an image space rendering we denote the appearance of the model at pixel $(x,y)$ by $I_{\text{model}}^{x,y}(\Theta)$. In the simplest case, the image space rendering is computed directly from the object space rendering using Gouraud interpolation shading:
\begin{equation}
    I_{\text{model}}^{x,y}(\Theta) = \mathbf{a}_{\mathcal{C},\mathcal{T},\mathbf{w}^s,\mathbf{w}^e}(x,y)^T
    \begin{bmatrix}
    I_{\text{model}}^{t^1_j}(\Theta) \\
    I_{\text{model}}^{t^2_j}(\Theta) \\
    I_{\text{model}}^{t^3_j}(\Theta) \\
    \end{bmatrix},
\end{equation}
where $j=\textbf{raster}_{\mathcal{C},\mathcal{T},\mathbf{w}^s,\mathbf{w}^e}(x,y)$. Other rasterization strategies may be more complex. For example, \citet{genova2018unsupervised} use rasterization in a differentiable deferred shading renderer more akin to Phong interpolation shading. Here, vertex normals and colors are rasterized and interpolated, then reflectance calculations are done in image space.

Note that overcoming the non-differentiable nature of rasterization is an open problem. \citet{kato2018renderer} present an approximately differentiable renderer based on rasterization. \citet{liu2019softras} propose a rasterizer in which triangles make a soft (and hence differentiable) contribution to image appearance. More ambitiously, differentiable rendering using other pipelines is now also being considered, for example, differentiable path tracing \cite{li2018differentiable}.

Very recently, the explicit fixed models used in conventional rendering are being augmented or replaced by learning components, so-called \emph{neural rendering}. For example, \citet{kim2018DeepVideo} train an image to image network that transforms a low-quality 3DMM rendering into a photorealistic video frame. 





\subsection{Open challenges}\label{sec:synthesis:open}

The image formation models used in the context of 3DMMs are much simpler than those used in graphics and many other areas of computer vision. For example, we are not aware of any work that allows for a center of projection different to the center of the image, even though many face image datasets consist of images cropped (probably non-centrally) from larger images. Similarly, nonlinear distortion is always ignored. The effect of this assumption is not understood. In other fields like structure-from-motion, it is standard to impose constraints derived from metadata, knowledge of physical camera parameters and so on. This is not currently being done to a significant extent with 3DMMs.

Advances in rendering in computer graphics are slowly propagated into the world of 3DMMs and especially into the analysis-by-synthesis process. One of the reasons is that almost every model extension makes the model adaptation more complicated and a lot of methods rely on the rendering process to be differentiable. There is a dramatic gap between what current computer graphics or also deep learning-based image generation methods are capable of and what is state of the art for 3DMMs. Also generated instances usually lack facial details like wrinkles or moles which are challenging to render properly. Recent work aims at those challenges by using generative adversarial networks as texture models \cite{slossberg2018high} but they are not modeled in the shape and not specially treated during rendering. A possible future direction is to either model or learn the gap between current 3DMM renderings and state of the art computer graphics or real-world 2D images.

An interesting open challenge is to better exploit the constraint of the 3DMM. Existing work uses generic pipelines for tasks such as rasterization or visibility calculation. However, the geometry is defined by a low dimensional parameter vector from which the per-vertex visibility could presumably be inferred more efficiently than treating the resulting mesh as a generic shape. The attempt of \cite{schneider2017efficient} to learn the relationship between PRT coefficients and shape parameters is a first step in this direction.

\section{Analysis-by-Synthesis}\label{sec:AbyS}

3DMMs have been widely used for image-based reconstruction. 
Reconstructing a 3D face from an observed image(s) involves estimating the 3DMM coefficients which can best explain the observation. 
This is the inverse of the image synthesis process covered in the previous section. 

Analysis-by-synthesis refers to a class of optimization problems which solves this by minimizing the difference between the observed image(s) and the synthesis of an estimated 3D face. 
Such an optimization problem can be ill-posed with several ambiguities and multiple minima.
This is a widely researched problem, with a variety of solutions exploring different input modalities (Sec.~\ref{Sec:Analysis-by-synthesis:input}),  energy functions (Sec.~\ref{Sec:Analysis-by-synthesis:energy}) and optimization strategies (Sec.~\ref{Sec:Analysis-by-synthesis:optimization}). We present publicly available approaches in Table~\ref{table:abyssoftware}.
\begin{table*}[h!]
\begin{tabular}{m{0.17\linewidth} m{0.15\linewidth} m{0.18\linewidth} m{0.15\linewidth} m{0.15\linewidth}} 
\toprule
\textbf{publication} & \textbf{input} & \textbf{estimates} & \textbf{approach} & \textbf{comment} \\ \midrule
Edge fitting \newline \cite{bas2016fitting} & 2D image, landmarks & pose, shape & edge features, ICP & \\ \hline
Eos fitting library\newline \cite{Huber2016} & 2D image, landmarks & pose, shape & landmark and contour fitting & \citet{huber2018phd} handles expressions \\ \hline
Basel Face Pipeline \newline\cite{bfm17} & 2D image, landmarks & pose, shape, expression, texture,  illumination & MCMC Sampling & estimates posterior distribution, \citet{Egger2018IJCV} handles occlusion\\ \hline
Deep 3D Face Reconstruction \newline\cite{deng2019accurate} & 2D image(s) & pose, shape, expression, texture, illumination & deep (ResNet)& \\ \hline
PRNet \newline \cite{feng2018prn} & 2D image & pose, shape & deep (convolutional) & outputs mesh in BFM topology\\ \hline
Expression-Net \newline\cite{chang2018expnet} & 2D image & pose, shape, expression, texture & deep (ResNet) & bundles \cite{chang2017faceposenet,tuan2017regressing}  \\ \hline
RingNet \newline\cite{RingNet:CVPR:2019} & 2D image & pose, shape, expression & deep (ResNet)& handles occlusion \\ \hline
Pix2vertex \newline\cite{SelaRK17} & 2D image & pose, shape, expression & deep + shape from shading& shape beyond 3DMM\\ \hline
Facial Details Synthesis \newline \cite{chen2019photo} & 2D image & pose, shape, expression, appearance & UNet for details &\\ \hline
3DMMs as STNs \newline \cite{bas20173d} & 2D image & pose, shape, expression & spatial transformer network & \\ \hline
3D Face Reconstruction \cite{tran2018extreme} & 2D image, output of \cite{tuan2017regressing} & shape details& estimate bump map using encoder-decoder architecture& handles occlusions \\ \hline
FLAME \newline\cite{FLAME2017} & 2D / 3D landmarks & pose, shape, expression &  landmark fitting  & \\ \hline
Basel Face Pipeline \newline\cite{bfm17} & 3D scan, landmarks & pose, shape, expression, texture & Gaussian process regression, nonrgid ICP & \\ \hline
LSFM Pipeline \newline \cite{LSFM2016} & 3D scan & pose, shape, expression & nonrigid ICP & fully automatic \\ \hline
Model Fitting \newline \cite{Brunton2014Review} & 3D scan, landmarks & pose, shape & nonrigid ICP, template and model fitting & handles occlusions\\ \hline
Multilinear Model Fitting \cite{BolkartWuhrer2015,Brunton2014} & 3D scan, landmarks & pose, shape, expression & nonrigid ICP, global model in \citet{BolkartWuhrer2015}, local model in \citet{Brunton2014} & handles occlusions \\

\bottomrule
\end{tabular}
\caption{Overview of publicly available model adaptation and registration frameworks for 3DMMs.}
\label{table:abyssoftware}
\end{table*}

%
%
Analysis-by-synthesis techniques have also recently been used in combination with deep learning architectures for learning-based reconstruction algorithms. We will discuss these methods in Sec.~\ref{sec:deeplearning}.
%

\subsection{Input Modalities}
\label{Sec:Analysis-by-synthesis:input}
Analysis-by-synthesis methods have been explored using multiple image modalities, from multi-view to monocular images and videos.
While multi-view methods produce very detailed and high-quality results, capturing such data requires expensive setups. 
A lot of recent focus has been on obtaining similar quality reconstructions with much lower cost solutions, e.g., using a single RGB image. 
This has also led to an increase in commercial applications for the mass market. 
Fitting a 3DMM to 3D scans can also be considered as analysis-by-synthesis. This is related to registration techniques, covered in Sec.~\ref{sec:modelling}.
\paragraph{Multi-View Systems}
We will start our discussion with multi-view solutions which minimize the photometric consistency between the multi-view images and the synthesis of the estimated reconstruction. 
%
%
%
%
%
%
%
Most multi-view methods, such as those covered in Sec.~\ref{sec:capture} do not require a strong prior in the form of 3DMMs. 
However, there are several methods which use 3DMMs to aid reconstruction in stereo camera systems.  
Model-based stereo reconstruction was explored in \citet{Wallraven99}. The reconstruction quality was improved by eliminating the estimation of illumination and reflectance in \citet{fransens2005parametric, amberg2007reconstructing}.  3DMMs also prove to be very valuable in low-resolution settings where high-quality image textures cannot be exploited, or under occlusions \cite{romeiro2007model, thies2018HeadOn}. 
Most of the methods discussed here solve very large optimization problems, and are not real-time. \citet{Thies2018facevr} is one real-time method which has a data-parallel implementation on a GPU.  
\paragraph{Monocular RGBD} RGB-D sensors capture RGB as well as depth information of the scene. 
Consumer stereo cameras either use passive stereo, IR projection-mapping, or time-of-flight technology.
The depth channel in the input helps in resolving depth ambiguities due to the lack of multiple views. 
Thus, in addition to photometric consistency, these methods also minimize depth consistencies using point-to-point and point-to-plane distances, see Sec.~\ref{sec:modelling}.
Since monocular reconstruction methods solve a smaller optimization problem compared to multi-view methods, many real-time solutions exist \cite{Bouaziz13, Li13, Thies15, Weise2011, Hsieh2015}.
While most methods heavily rely on 3DMMs, some try to adapt them to capture user-specific details. 
\cite{Weise2011} build a user-specific expression model by adapting a general one. This is done in an offline stage before the online tracking. 
\cite{Bouaziz13, Li13} adapt the 3DMM online, thus removing the need for an offline step.  
\cite{Hsieh2015} introduced an occlusion robust tracking system using face segmentation masks.
\cite{Liang2014} reconstruct a single image by retrieving instances of 3D shapes from a dataset and merging them, thus avoiding the need for 3DMMs. 
\paragraph{Monocular RGB} Without the presence of the depth channel, the analysis-by-synthesis problem becomes even more ill-posed. 
These methods cannot easily resolve depth ambiguities. Thus, the prior knowledge of a 3DMM becomes important. 
Monocular RGB videos can provide more constraints. The identity component, in this case, can be estimated by fusing information from multiple frames in a preprocessing step.
Many methods can track the face in real-time \cite{Thies16, Ichim2015, cao2016, Cao2015, Cao2014, Cao2013}. 
As in the case of RGB-D based methods, there are methods which try to add details over the 3DMM reconstructions to make the results user-specific and detailed. 
\cite{garrido2016corrective} add medium-scale correctives based on spectral basis vectors. \citet{garrido2013reconstructing,garrido2016corrective, Cao2015, Suwajanakorn2014, shi2014automatic} also add high-frequency wrinkle-level details. \citet{Wu2016} use local blendshape models to capture more details compared to global blendshape based methods. 
\cite{Ichim2015, Cao2013, cao2016} compute user-specific 3DMMs using images of a person performing specific known expressions. 

Photo-collections, i.e., collections of images of a person can also be used to constrain the identity components of the reconstructions \cite{liang2016head, Kemelmacher-Shlizerman2011, suwajanakorn2015makes, Roth2015, Roth2016, Piotraschke2016}. This is a more unconstrained setting compared to multi-view images where all views are captured at the same time in the same environment. Approaches which use photo-collections and videos are more practical than multi-view images since such data is widely available for most people.

\begin{figure}
    \centering
    \includegraphics[width=\linewidth]{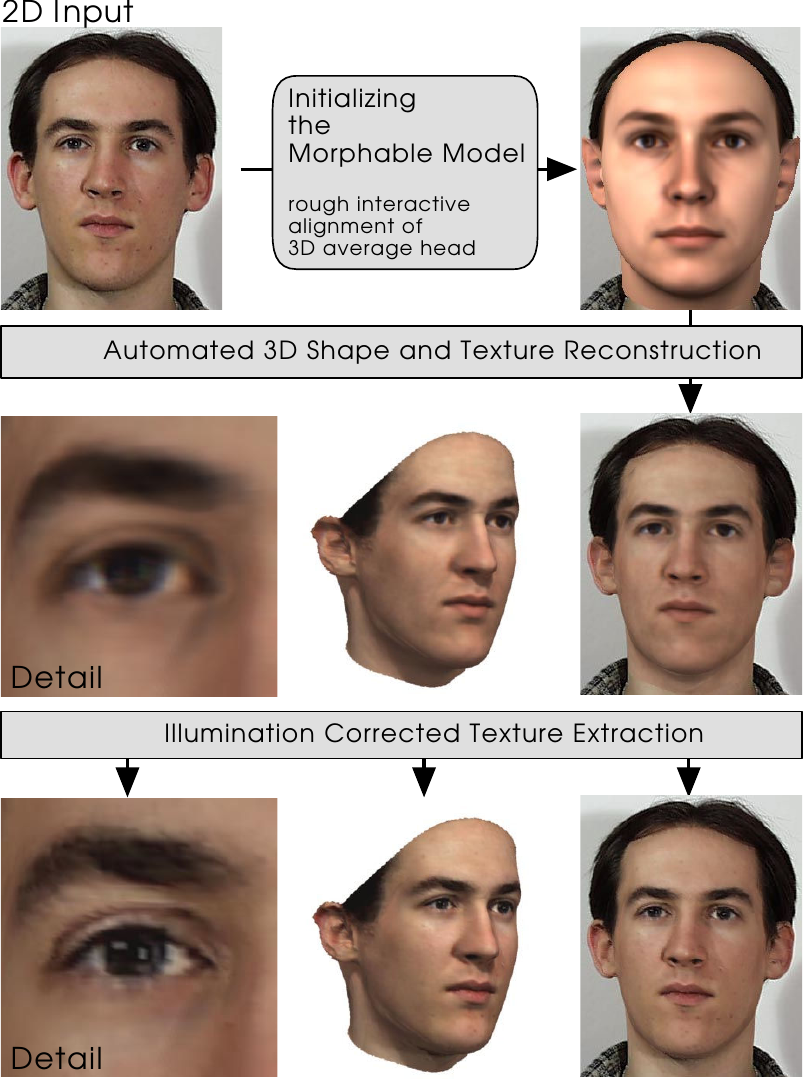}
    \caption{The analysis-by-synthesis pipeline used by \citet{BlanzVetter1999} for reconstruction from a single image. The different steps include initialization, optimization, and refinement of the optimized 3DMM texture.}
    \label{fig:A-by-S}
\end{figure}

Reconstruction from a single image is the most challenging scenario. However, the original work of \citet{BlanzVetter1999} already proposed an analysis-by-synthesis solution, see Fig.~\ref{fig:A-by-S}. 
While they required manual initialization for the optimization problem, several approaches made the approach more robust to enable automatic reconstruction \cite{ Tewari2018, fried2016perspective, paysan2009face, Schoenborn2017, Egger2018IJCV, kortylewski2018informed, schneider2017efficient, bas2016fitting, Hu2017, aldrian2011inverse}.
Most analysis-by-synthesis approaches evaluate the photometric consistency between the observations and the estimates. Some approaches have explored the use of other image features, such as edges, or SIFT \cite{Romdhani2005, booth20173d}, in order to obtain higher fidelity reconstructions.
Occlusion robust reconstruction by jointly solving for segmentation has been explored in \citet{Egger2018IJCV}.
Monocular reconstruction methods primarily differ in their formulated energy functions. We will look at these in detail in Sec.~\ref{Sec:Analysis-by-synthesis:energy}.
\subsection{Energy Functions}
\label{Sec:Analysis-by-synthesis:energy}

The analysis-by-synthesis paradigm involves the solution of a nonlinear optimization problem made up of a number of energy functions. Methods differ in their combination and precise design of these energy functions, their relative weights and (dealt with in the following subsection) the optimization strategy used to minimize the energy. Here we describe the most commonly used energy functions for fitting to RGB images.
For single image reconstruction, the energy functions are expressed in terms of a single set of unknown parameters $\Theta$. In the case of multi-view images of a static face, the camera parameters are indexed by viewpoint while all others are fixed across views. In the case of an image sequence of a dynamic face, camera, expression  and lighting parameters are indexed by frame while neutral shape and texture parameters are fixed throughout the sequence.

\paragraph{Appearance error} The key ingredient of analysis-by-synthesis is to measure the difference between observed data and a synthesis using the model. Most directly, this is the appearance error between an input image and the rendered face. A number of variants of this term have been used. The \emph{pixel-wise} formulation sums the appearance error over the pixels of the image, necessitating rasterization of the model:
\begin{equation}
\label{eq:pixel_appearance}
E^{\text{pixel}}_{\text{appearance}}(\Theta, I_{\text{obs}}) =   \sum_{(x,y)\in\textbf{foreground}} \left\| I_{\text{obs}}(x,y) - I_{\text{model}}^{x,y}(\Theta) \right\|^2
\end{equation}
where $\textbf{foreground}=\{(x,y)\in\mathcal{I}|\textbf{raster}_{\mathcal{C},\mathcal{T},\mathbf{w}^s,\mathbf{w}^e}(x,y)\neq\texttt{NULL}\}$ is the set of pixels covered by the union of all triangles. This formulation naturally weights the contribution of model vertices in terms of their contribution to the appearance of a pixel. An alternative is to compute the appearance error \emph{vertex-wise} by rendering in model space and sampling the image intensities onto the vertices:
\begin{multline}
E^{\text{vertex}}_{\text{appearance}}(\Theta, I_{\text{obs}}) =  \\  \sum_{j\in\textbf{visible}} \left\| \textbf{interp}[I_{\text{obs}},\textbf{project}[\mathcal{C},\mathbf{v}_j(\mathbf{w}^s,\mathbf{w}^e)]]-I_{\text{model}}^j(\Theta) \right\|^2
\end{multline}
where $\textbf{interp}[X,(x,y)]$ represents differentiable interpolation of 2D object $X$ at location $(x,y)$ and $\textbf{visible}$ is the set of visible vertices. A common variant of this approach uses a random subset of vertices rather than all of them. This is more efficient, introduces stochasticity that may help avoid local minima and avoids overly conservative fits near the boundary where background may be sampled. Differentiable interpolation of the image can either be done with explicit differentiable sampling (e.g.,~bilinear sampling as in \citet{jaderberg2015spatial}) or precomputing the image gradient and then interpolating this along with the image intensities (this was done in the original \citet{BlanzVetter1999} paper). A drawback of the vertex-wise error is that regions of the image with dense coverage from projected vertices are weighted more heavily than more sparsely sampled regions. This can be overcome by using weights related to projected area. \citet{BlanzVetter1999} accomplished the same effect by using the triangle area as the probability by which the triangle would be selected in their random sampling.
For multi-image methods, the above energies are simply summed over each image.

\paragraph{Feature-based energies}

There are other, less direct, ways to compute an error between the observed data and model. This is done by first computing features from observed data and then measuring the difference between those features and the corresponding ones in the model. By far the most commonly used features are landmarks (alternatively known as keypoints or fiducial points) which often used for initialization and are still important in much of the state-of-the-art, e.g., \cite{RingNet:CVPR:2019}. 
A landmark detector returns a set of 2D landmark coordinates $\{\mathbf{x}_j\}_{j=1}^J$ with $\mathbf{x}_j\in\R^2$. As a one-off procedure, each landmark is associated with the corresponding vertex in the 3DMM such that the $j$th landmark corresponds to vertex index $k_j\in\{1,\dots,n\}$. The reprojection error of the model landmarks with respect to detected positions is then given by:
\begin{equation}
    E_{\text{landmarks}}(\Theta, \{\mathbf{x}_j\}_{j=1}^J) = \sum_{j=1}^J \left\| \mathbf{x}_j - \textbf{project}[\mathcal{C},\mathbf{v}_{k_j}(\mathbf{w}^s,\mathbf{w}^e)] \right\|^2.
\end{equation}
Sometimes the landmarks are allowed to slide on the face surface such that each landmark has a set of vertices to which it could correspond \cite{Zhu:15}.

Edges directly convey geometric information about occluding boundaries and texture edges. Misalignments between model and image edges seriously degrade the perceptual quality of a reconstruction and lead to the wrong part of the face, or the background, being sampled onto the mesh. \citet{moghaddam2003model} were the first to exploit this cue by fitting to multi-view silhouettes. \citet{Romdhani2005} computed the distance transform of detected edges in an input image providing a distance-to-edge cost surface that was sampled at projected positions of vertices lying on model texture edges or the occluding boundary. \citet{amberg2007reconstructing} extended this to multiple views and improved robustness by averaging the cost surface over different parameters of the edge detector. \citet{keller20073d} showed that these cost functions are neither continuous nor differentiable. \citet{bas2016fitting} transformed edge fitting into landmark fitting by alternating between computing an explicit correspondence between edge pixels and model edges and minimizing the resulting landmark energy. \citet{sanchez2016statistical} directly regress shape parameters from a set of multi-view occluding contours.

Finally, some other features have been considered. \citet{Romdhani2005} used the position of specularities in the image to constrain the surface normal direction at the corresponding location on the model via a specular reflection model. \citet{booth20173d} and \citet{booth20183d} compute dense SIFT features from the input image and compare these to the SIFT features on which their statistical texture model is built in a similar fashion to the vertex-wise appearance error above. 

\paragraph{Background Modeling}
A common challenge when optimizing for pose and shape is the varying visibility of vertices for \emph{vertex-wise} errors and the varying number of pixels covered by the face for \emph{pixel-wise} errors. This leads commonly to the undesired effect of shrinking. Having the model covering fewer pixels or having fewer vertices visible leads to an undesired local optimum of most error terms. Common strategies to overcome this are fixed visibility, restrictive regularization, relying on landmarks, enforcing edge or contour terms or explicit image segmentation. \citet{schonborn2015background} demonstrated the problems with an implicit background model which is present in all error formulations and have shown that even simple background models $b$ like a constant, a Gaussian or an image histogram-based model can solve this issue.
The background model can easily be added to the existing formulations, e.g.,~for the \emph{pixel-wise} formulation as:
\begin{multline}
E^{\text{image}}_{\text{appearance}}(\Theta, I_{\text{obs}}) = \\ E^{\text{pixel}}_{\text{appearance}}(\Theta, I_{\text{obs}}) +  \sum_{(x,y)\in\textbf{background}} b(I_{\text{obs}}(x,y)).
\end{multline}

\paragraph{Occlusions and Segmentation}
Occlusion of faces by other objects, that are not part of the generative model, are a common challenge for the so far presented error terms and for analysis-by-synthesis in general. There are various methods presented on how to identify occlusions. Those methods range from appearance-based methods \cite{de2006generalized, pierrard2008skin} to detection \cite{morel2016generative} and segmentation-based methods \cite{Saito2016, Egger2018IJCV}. They share the basic idea, that occluded pixels are excluded from the model evaluation:
\begin{equation}
E^{\text{semantic}}_{\text{appearance}}(\Theta, I_{\text{obs}}) =  
\sum_{l \in \textbf{label}}\sum_{(x,y) \in R(l)} E^{\text{pixel}}_{\text{label}}(\Theta, I_{\text{obs}}(x,y), l),
\end{equation}
%
where each $E^{\text{pixel}}_{\text{label}}$ is a separate model per label, and $R(l)$ is the image region covered by label $l$. Those labels could e.g., be face, occlusion and background or also contain more detailed labels like beards.
 Whilst the segmentation is based on detection and fixed in \citet{pierrard2008skin, morel2016generative, Saito2016}, other methods solve for segmentation and model parameter estimation jointly in an Expectation-Maximization-based manner \cite{de2006generalized, Egger2018IJCV}.


\paragraph{Priors}
A 3DMM is a statistical model and so provides a natural probabilistic prior over the parameter space. Under the assumption that the original data is Gaussian distributed the natural cost function to express this prior for either the shape or texture model is:
\begin{equation}
    E_{\text{prior}}(\Theta) = \sum_{i=1}^d \frac{w_i^2}{\sigma_i^2},
\end{equation}
where $\sigma_i^2$ is the variance associated with the $i$th principal component. The drawback to this prior is that it is minimized by the mean face and, if weighted heavily, leads to model dominance where recovered faces are too close to the average. There has been in discussion in the literature \cite{patel2016manifold,lewis2014probable} as to whether this prior is appropriate in high dimensional space and alternatives have been considered, as will be described next.

One class of techniques allows reconstructed shape and/or texture to deviate from the 3DMM subspace enabling recovery of fine-scale detail not captured by the model. Allowing arbitrary shape or albedo changes transforms the problem into classical shape-from-shading and becomes highly ill-posed. For this reason, additional generic priors are used. \citet{patel2012driving} use a piecewise smoothness prior on per-vertex diffuse albedo which is allowed to vary per-vertex along with surface normals to satisfy a shape-from-shading constraint. This is regularized using the squared vertex distance between the updated shape and the closest shape in the 3DMM space. \citet{richardson2017learning} use the same regularization, though, expressed in terms of per-pixel depth. To ensure smoothness, they also use the L1 norm of the discrete Laplacian of the depth map. The L2 norm of the mesh Laplacian has also been used as a smoothness prior \cite{garrido2016reconstruction, Tewari2018}

When reconstructing a dynamic face from video, parameters can either be assumed fixed (if identity dependent) or smoothly varying (pose, expression, lighting). These latter parameters can, therefore, be regularized with generic temporal smoothness priors. A common and simple way to express this prior is to initialize each frame with the estimate from the previous one. This encourages convergence to a local minimum close to the solution for the previous frame. More sophisticated priors have also been considered. For example, \citet{Cao2013,Weise2011} build a Gaussian mixture model over expression parameters from the previous $k$ frames. This model is then used to regularize the estimate for the current frame. 

%

\subsection{Optimization}
\label{Sec:Analysis-by-synthesis:optimization}

From the perspective of optimizing the energy functions above, there are a number of significant challenges. First, most of the energy terms are nonconvex in theory and we observe in practice that there are many local minima. Second, the appearance error is not even continuous due to rasterization/vertex visibility and shadowing all being noncontinuous functions. Third, the appearance error has a small basin of convergence. When a model feature is completely misaligned to the image (or in the extreme case, the whole model aligned entirely to background), the gradient of the appearance error conveys no useful information. 
Fourth, all parameters have global influence. 
Fifth, computing the appearance error and its gradient is computationally expensive, amounting to the rendering of an image. For these reasons, a significant effort has gone into the selection of optimization algorithms and engineering of the optimization schedule to develop methods that are sufficiently fast and robust.


The majority of existing approaches optimize based on gradient information of the energy function. The original \citet{BlanzVetter1999} approach used first-order gradient descent, as have other more recent methods \cite{fried2016perspective, Ichim2015, Bouaziz13}. Since they computed the appearance error over only a small subset of randomly selected triangles, this is strictly \emph{stochastic} gradient descent (SGD). An interesting parallel here is that modern deep learning-based methods (see Section \ref{sec:deeplearning}) are usually trained with SGD and use similar energy functions so they are learning from the same signal used in the original method.

Since the energy terms above can easily be formulated as nonlinear least-squares problems, specialized pseudo-second-order methods like Gauss-Newton or Levenberg-Marquardt have often been used \cite{garrido2013reconstructing, garrido2016corrective, garrido2016reconstruction, Thies15, Thies16, Romdhani2005}. \citet{booth20173d} use a ``\emph{project-out}'' strategy in which appearance parameters are implicitly solved in a least squares sense and optimization takes place only over geometric parameters. General pseudo-second-order methods such as BFGS have been used \cite{Cao2013,Weise2011} as well as genuine second-order methods, specifically a stochastic variant of Newton's method \cite{blanz2003face}. 
%
As the problem size increases, as in the case of shape-from-shading, gradient descent becomes the most common optimization approach \cite{garrido2016reconstruction, shi2014automatic, Tewari2018, Suwajanakorn2014}.
In all the above methods, the discontinuity of the appearance function is dealt with by fixing rasterization/visibility when computing gradients or even keeping them fixed for a certain number of iterations. Importantly, this means that the gradient cannot convey information about a change in visibility. Many other tricks have been considered, for example, hierarchical optimization (both in parameter space and spatially, i.e.~multiresolution \cite{Thies16}) and using an optimization schedule in which different energy functions are switched on or weighted differently at different phases on the optimization \cite{BlanzVetter1999}.

Several approaches have decomposed the energy terms into several smaller, often linear, problems (sometimes with closed-form solutions) that can be solved efficiently and in sequence
\cite{romdhani2002face,hu2017efficient,aldrian2013inverse,aldrian2010linear,aldrian2011inverse,aldrian2011inverseb,Cao2013,Cao2014,Zhu:15,Saito2016,bas2016fitting}. These alternating approaches are usually very efficient but not guaranteed to obtain the optimum solution that comes from optimizing all parameters simultaneously.

Gradient-based methods are typically initialized by fitting only to landmarks, i.e.~to optimize the landmark energy in isolation. 
Originally, the landmark positions were provided manually but combining with an automatic landmark detector provided fully automatic methods \cite{breuer2008automatic}. From a landmark detector that outputs many hypothesized landmark locations, including many false positives, \citet{amberg2011optimal} use Branch and Bound to select the subset configuration of landmarks that is most consistent with the 3DMM. \citet{bas2017does} show how to express the landmark energy as a separable nonlinear least squares problem. 


While gradient-based methods are widely used mainly due to computational efficiency and ease of implementation, these methods are sensitive to initialization and often end up in local minima.
Probabilistic methods based on Bayesian inference were proposed to deal with these limitations \cite{Schoenborn2017, Egger2018IJCV,  kortylewski2018informed, schneider2017efficient}. These methods do not require any gradient computation of the energy terms to update the estimates.  They are stochastic and thus, less susceptible to getting stuck in local minima.  Different from optimization-based methods which only provide a single solution, these approaches approximate the full posterior distribution and thus provide access to a manifold of possible solutions.



\begin{figure}
    \centering
    \includegraphics[width=\linewidth]{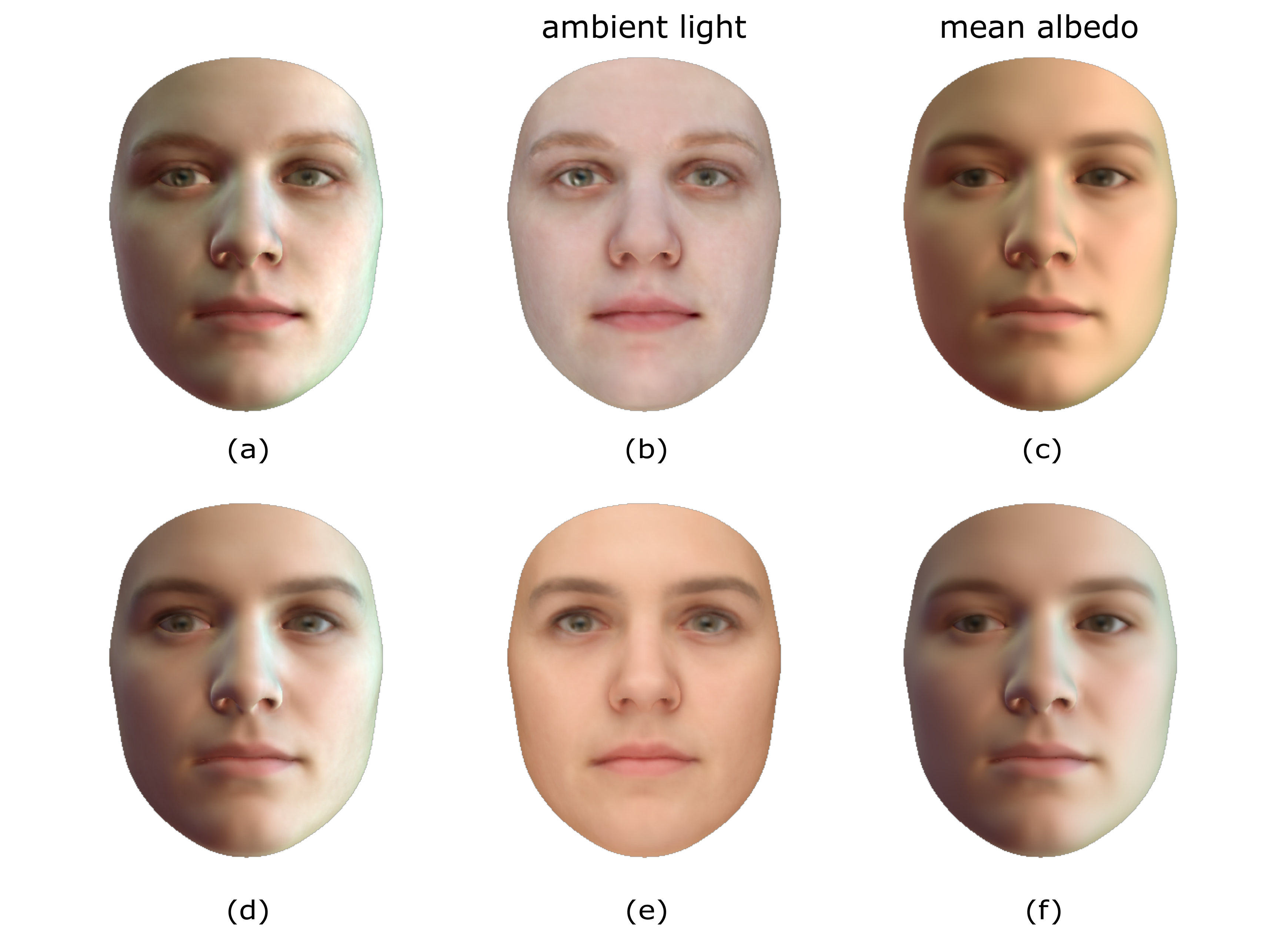}
    \caption{An example of the albedo-illumination ambiguity presented by \citet{egger17phd}. The target image in the first row (a), its color rendered under ambient illumination (b) and its illumination rendered on the mean albedo of the Basel Face Model (c). The second row shows a model instance with different color (e) and illumination (f) parameters but very similar appearance (d).}
    \label{fig:ambiguity}
\end{figure}

\subsection{Open challenges}
Reconstructing 3D shape and albedo from a 2D image is an ill-posed problem. Ambiguities like the perspective face shape ambiguity \cite{smith2016perspective} and the albedo illumination ambiguity \cite{egger17phd} have been demonstrated (see Figure~\ref{fig:ambiguity}).  These ambiguities can not be resolved completely and priors are our best approach to at least find a reasonable estimation.  They are the major reason why there is a huge gap between the estimates we can get from multi-view and 3D data vs. from monocular images. Even in the state-of-the-art, it is often evident that overall skin color is explained using the lighting while the albedo colors are similar for very different skin types \cite{tewari2018self}. This is somewhat improved by discriminative methods that do not need to synthesize the same appearance as a given image, only an image with the same identity \cite{genova2018unsupervised}, thereby sidestepping explicit estimation of illumination and camera parameters. Reporting the geometric errors obtained by the model mean is not common. Only three papers demonstrated their 3D reconstructions to be closer in mesh distance to the ground truth face compared to the model mean \cite{aldrian2013inverse, Schoenborn2017,RingNet:CVPR:2019}.  

Current state of the art techniques also lack dramatically in accuracy across pose and in terms of matching the contours and edges. It is very difficult to evaluate these beyond qualitative evaluation which makes it difficult to compare different approaches. Recently a first benchmark with natural images and ground truth shape was published and will help to better compare competing methods \cite{RingNet:CVPR:2019}. However, as methods get more accurate, the mesh distance errors get close to the range of error in computing  ``ground truth'' using multi-view methods. This makes it difficult to quantitatively compare different approaches.

Another challenge which is usually neglected are occlusions. Faces are mostly occluded by objects which are frequently in front of faces like glasses, cigarettes, hands or microphones, but can also be occluded by virtually every other object. Analysis-by-synthesis methods fail when they do not explicitly model occlusions.
Furthermore, reconstruction methods based on 3DMMs are limited to the space of faces covered by the models. A lot of residual error in the results stems from the fact that 3DMMs do not model detailed and high-frequency geometry and texture.  Furthermore, most approaches use simple lighting models which cannot explain many in-the-wild images. These limitations are also shared with the learning-based methods which use analysis-by-synthesis in their pipeline, see Sec.~\ref{sec:deeplearning}. 

Recent techniques have been focused on reconstruction from a single face individually. The aim of face image analysis would, however, go beyond interpreting a single face or each face separately. We would like to analyze and interpret interactions between people and perhaps also ease the analysis task by exploiting scene constraints, such as shared illumination parameters to deal with albedo-illumination ambiguity, or constraints on the perspective face shape ambiguity by analyzing multiple faces jointly.


\section{Deep Learning}
\label{sec:deeplearning}

\begin{figure*}
    \centering
    \includegraphics[width=\textwidth]{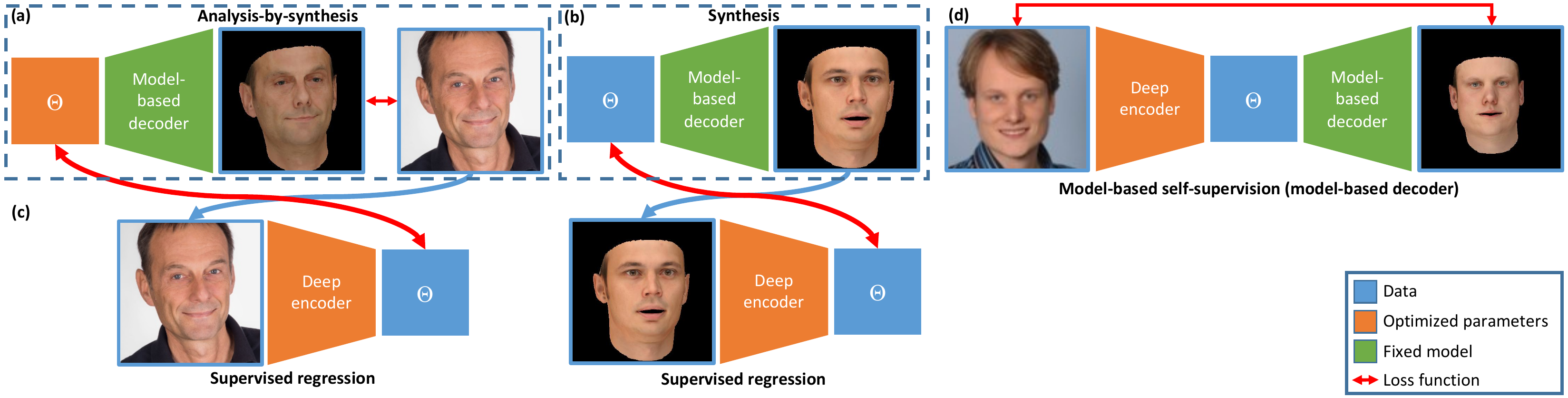}
    \caption{The relationship between classical analysis-by-synthesis and deep learning approaches. (a) analysis-by-synthesis, e.g.,~\cite{BlanzVetter1999}. (a)+(c) training a regressor based on the output of an analysis-by-synthesis algorithm, e.g.,~\cite{tuan2017regressing}, (b)+(c) training a regressor using synthetic data generated by a model, e.g.,~\cite{richardson20163d}, (d) self-supervision, e.g.,~\cite{tewari17MoFA}.}
    \label{fig:architectures}
\end{figure*}

So far, we have mainly discussed classical face modeling and parameter estimation techniques based on optimization-based inverse graphics. We now discuss how these processes can be replaced by or combined with deep learning, see Figure~\ref{fig:architectures}. There are a number of reasons for wanting to do this. On the modeling side, the use of nonlinear, deep representations offers the possibility to surpass classical linear or multilinear models in terms of generalization, compactness and specificity \cite{Styner03}. On the parameter estimation side, we can exploit the speed and robustness of deep networks to achieve reliable performance on uncontrolled images.

We begin by discussing deep modeling and deep model fitting before finally discussing methods that simultaneously learn both the model and how to fit it within a single deep network.

\subsection{Deep Face Models}

The traditional modeling techniques discussed in Section~\ref{sec:modelling} aim to represent face shape, expression, and appearance as vector $\vec{w}$ in a low-dimensional latent space $\mathbb{R}^d$. The projection into (respectively reconstruction from) this latent space is defined by linear or multi-linear operations, and can be thought of as encoding (respectively decoding) the high-dimensional information in $\mathbb{R}^d$. Deep learning provides a new tool for building 3DMMs using nonlinearities both in the encoder and the decoder. This way of building morphable models is currently a very active area of research. 

We can see the relationship between the encoder and decoder learned using deep learning and classical works using the example of linear models commonly used for shape and texture modeling. In the context of deep learning, such a linear model formalized in Equation~\eqref{eq:pca}, is exactly equivalent to a fully connected layer in a neural network. Concretely, the parameter vector $\mathbf{w}$ plays the role of the input features, the principal components $\vec{e}_j$ are the weights and the mean $\bar{\vec{c}}$ is the bias. This can be viewed as \emph{decoding} from the latent parameter space to the data space $\mathbf{c}$. Projection onto the model can similarly be viewed as \emph{encoding} with a fully connected layer in which the input features are the data, the weights are the rows of the transposed principal component matrix and the biases are given by $-\vec{e}_j^T\bar{\vec{c}}$. Concluding the analogy, a PCA can be accomplished by combining the encoder and decoder as a linear autoencoder with a single hidden layer. Such an autoencoder with $d$ neurons in the hidden layer will learn a latent space with the same span as a $d$ dimensional PCA, though without the guarantee of orthogonality (though this could be ensured with appropriate loss functions).

Given this close relationship between classical methods and deep learning, it is natural to ask if there exist more powerful nonlinear models that can be trained based on current advances in deep neural networks. As in the classical work, this has been considered for the 2D case. \citet{Duong2019} propose a deep appearance model for 2D facial images that extends 2D AAMs to model nonlinearities. This is achieved using deep Boltzmann machines to model 2D shape and texture information.  For modeling 3D faces, first successful models using autoencoders, GANs, and hybrid structures have been proposed, as detailed in the following.

\citet{Abrevaya18} proposed the first encoder-decoder architecture to model the 3D geometry of faces. The encoder first projects the 3D face to a 2D image and uses a standard image-based encoder, while the decoder is fixed to a classical tensor-based face model. This allows decoupling shape variations caused by identity and expression. \citet{Bagautdinov2018} introduced a VAE that models different levels of detail of facial geometry by representing global and increasingly localized shape variations in different layers of the network. The 3D geometry is again represented using a two-dimensional mapping, and convolutions are performed in the image domain. This work allows representing highly detailed geometric information in latent space. \citet{Lombardi2018deep} extend this work to jointly encode variations in appearance and geometry, for the application of highly detailed facial rendering from novel viewpoints. \citet{CoMA2018} proposed the first autoencoder architecture for the geometry of faces that performs convolutions in 3D mesh space directly instead of going through a 2D image representation. The model, named CoMA, allows for very compact representations of the facial geometry. This work was recently extended to encode both texture and shape information jointly~\cite{Zhou2019}.

An alternative line of work considers learning GANs for 3D face modeling. \citet{slossberg2018high} proposed the first 3DMM using GANs. In this work, the facial texture is mapped to a coherent 2D image domain, and two-dimensional convolutions are employed to build a GAN of facial texture. This is combined with a standard PCA-based 3DMM for facial geometry, where for a generated face texture, a suitable PCA-based geometry is computed.  Recently, multiple methods were proposed to generate 3D facial geometry, possibly with texture information. \citet{Fernandez2019} proposed to train a GAN for the geometry of 3D faces that is able to decouple different factors of variation such as identity and expression. \citet{Shamai2019} proposed a GAN architecture to generate both facial geometry and texture, with a focus on highly detailed texture information by mapping the face to a unit rectangle. \citet{Cheng2019} proposed the first intrinsic GAN architecture that operates directly on 3D meshes. As in the case of 2D images, GANs are generally able to generate more detailed and realistic 3D faces than autoencoders at the cost of being more difficult to train.

Finally, hybrid structures can be effective to learn nonlinear 3DMMs. \citet{Tran2018} jointly learn a 3DMM and 3D reconstruction from a 2D image using a differentiable renderer in the training loss, see also Section~\ref{sec:3Dfrom2D}. The network takes as input a 2D image and encodes it into projection, shape and texture parameters. Two decoders are then used to infer 3D shape and texture, respectively. \citet{wang2019} proposed an adversarial auto-encoder structure that allows disentangling factors of variation such as identity, expression, or pose of 2D facial images, and that is trained in an unsupervised way. While the method's input and output are 2D images, the 3D geometry of the face can be reconstructed.

Recently, appearance modeling approaches based on deep learning have also been proposed. 
The rise of deep learning methods facilitated to learn per-vertex appearance models directly from images, such as done by \citet{tewari2018self}, who learn per-vertex albedo model offsets in order to improve the generalization ability of an existing PCA-based model. Similarly, \citet{t19fml}, learn a per-vertex albedo model from scratch based on video data. 
\citet{Zhou2019} train a mesh decoder that jointly models the texture and shape on a per-vertex basis, which, however, relies on the availability of 3D shape and appearance data.
There are also several deep learning approaches that consider a texture-based appearance modelling. Without the need of 3D data, \citet{Tran2018} learn a nonlinear facial appearance model represented in \emph{uv}-space based on CNNs, which, however, does not explicitly consider lighting. In follow-up work, the authors considered a more elaborate model where the albedo and the lighting is separately modeled ~\cite{tran2018learning,tran2019towards}.
Moreover, a range of generative methods that synthesize facial textures have been proposed,~e.g., by \citet{saito2017photorealistic}, \citet{slossberg2018high}, \citet{deng2018uv}, \citet{Lombardi2018deep}, \citet{nagano2018pagan} and \citet{yamaguchi2018high}. 
\citet{gecer2019ganfit} use GAN-based texture model for the task of 3D face reconstruction, and \citet{Nagano2019} use GAN-based texture models for the task of face normalization.

\subsection{Deep Face Reconstruction}

In the following, we discuss dense monocular face reconstruction approaches that are based on deep neural networks.
We discuss requirements on the used training data, as well as different training strategies.
Let us first have a closer look at the reconstruction problem, \citet{BlanzVetter1999} tackle monocular face reconstruction by fitting a parametric model based on an optimization approach, i.e., gradient descent.
Deep learning approaches follow a similar optimization strategy, but instead of solving the optimization problem at `test' time, they for example train a parameter regressor based on a large dataset of training images, see Figure~\ref{fig:architectures}.
The regressor can be interpreted as an encoder network that takes a 2D image as input and outputs the low-dimensional face representation.
Learned encoders can be combined with decoders based on classical face models to give rise to end-to-end encoder-decoder architectures.
This methodology is widely-used and enables the fusion of classical model-based and deep learning approaches.


\subsubsection{Supervised Reconstruction}
Supervised regression approaches are trained based on paired training data, i.e., a set of monocular images and the corresponding ground truth 3DMM parameters.
One of the essential questions here is how to efficiently obtain the ground truth for such a supervised learning task.
In the following, we will categorize the approaches based on the type of employed ground truth training data.

One option would be to let users annotate the ground truth.
While this is a popular strategy, which is often employed for sparse reconstruction problems \cite{Saragih11b}, the accurate annotation of dense geometry, appearance, and scene illumination is almost intractable.
A related approach is for example employed in the work of \citet{olszewski2016high}, where three professional animators manually created the blendshape animation to match a video clip.

For dense reconstruction tasks, some approaches \cite{Laine:2017} are trained based on images captured in a controlled multi-view capture setup.
Thus, ground truth can be obtained by a multi-view reconstruction approach followed by fitting a 3DMM to the resulting 3D data.
Normally, the ground truth is of very high quality, but the distribution of the captured monocular images does not match in-the-wild data, which can lead to generalization problems at test time.

The approach of \citet{tuan2017regressing} performs monocular reconstruction for multiple images of the same person and computes a consolidated face identity based on simple averaging of the 3DMM parameters.

Currently, many approaches \cite{richardson20163d,SelaRK17,McDonagh2016,klaudiny2017real,kim17InverseFaceNet,feng2018prn,yu2017learning} in the research community are trained on synthetic training data, since it is easy to acquire and comes by design with perfect annotations.
Given a face 3DMM, random identities and expressions can be sampled in parameter space.
Afterward, the models can be rendered under randomized illumination conditions and from different viewpoints to create the monocular images.
Often, background augmentation is employed by rendering the generated faces on top of a large variety of real-world background images. 
Since all the parameters are controlled, they are explicitly known and can be used as ground truth.
While it is easy to get access to synthetic training data, there is often a large domain gap between synthetic and real-world images, which severely impacts generalization to real images.
For example, hair, facial hair, torsos, or mouth interiors are often not modeled at all.
One possibility to counteract this problem in the future would be better models that include all these components.

To leverage the advantages of both real as well as synthetic training data, many current approaches \cite{richardson2017learning,kim17InverseFaceNet} are trained on a mixture of data from these two domains.
The hope here is that the approach learns to deal with real-world images, while the perfect ground truth of the synthetic training data can be used to stabilize training.
One interesting variant of this is self-supervised bootstrapping \cite{kim17InverseFaceNet} of the training corpus.
Other approaches that can be trained without requiring ground truth data are presented in the next sections.

\subsubsection{Self-Supervised Reconstruction}


Supervised training of a convolutional neural network requires an annotated dataset.
Most of the methods we have discussed so far use such datasets, either synthetic or real. 
Recently, some approaches explored self-supervised learning i.e., training on real image datasets without any 3D labels. 
This was made possible by a combination of analysis-by-synthesis (Sec. 4) and deep learning techniques. 
\citet{tewari17MoFA} introduced a model-based encoder-decoder architecture, which replaces the trainable decoder with an expert-designed fixed decoder.
This expert-designed decoder takes the 3DMM parameters (latent code) predicted by an encoder as input and transforms it into a 3D reconstruction using the 3DMM.
It further renders a synthetic image of the reconstruction using a differentiable renderer. Extrinsic parameters required for rendering are also predicted by the encoder. 
The loss function used is very similar to those used in analysis-by-synthesis (Sec.~\ref{Sec:Analysis-by-synthesis:energy}), consisting of photometric alignment and statistical regularization. 
We can think of such a technique as a joint analysis-by-synthesis optimization problem over a large training dataset, instead of a single image, see Figure~\ref{fig:architectures}.
This allows for training a parameter regressor without any 3D supervision.
This concept, usually in combination with supervised synthetic data has also been explored using higher-level loss functions like identity preservation \cite{genova2018unsupervised, RingNet:CVPR:2019}, or perceptual and adversarial losses \cite{tuan2017regressing}. 
\citet{gecer2019ganfit} employ GANs in combination with differentiable rendering to learn a powerful generator of facial texture.
\cite{sengupta2018sfsnet, richardson2017learning} refine 3DMM predictions for higher quality or more detailed results.
\cite{deng2019accurate, RingNet:CVPR:2019} extend the network architecture to allow for training using multiple images of a person as constraint. \citet{bas20173d} use a 3DMM as a spatial transformer network such that model fitting is learned as a by-product of solving a downstream task.

\subsection{Joint Learning of Model and Reconstruction}
\label{sec:3Dfrom2D}
Model-based encoder-decoder networks consist of a trainable encoder and a fixed decoder, where the decoder implements a 3DMM.
However, the 3DMM itself could be trainable. We could simply update its values using the gradients from the loss function. 
This would allow face model learning using only 2D supervision.
Learning 3D models entirely from 2D data was first shown in \cite{cashman2012shape} without the use of deep learning.
Several deep learning approaches have explored refining an existing 3DMM using large image datasets \cite{tewari2018self, tran2018learning, tran2019towards, Lin2020cvpr}. 
Nonlinear convolutional decoders have also be used to build nonlinear face models \cite{tran2018learning, tran2019towards}.
Models learned from 2D data are more generalizable to different identities, as the image datasets contain significantly more identities compared to the 3D datasets used to compute 3DMMs.  
Recently, an extension of the model-based encoder-decoder architecture was used to learn the identity component of a face model from videos \cite{t19fml}.

\subsection{Open Challenges}
Applying deep learning to the analysis of 3D face data is an active research topic that the community has only started to explore during the past few years, with many ongoing advances. Hence, many challenges currently remain to be solved. The most pressing ones include analyzing the limitations of current methods and providing comprehensive comparisons. This includes a clear analysis of the methods' tendency to overfit, especially when mostly synthetic data is used for training and the interpretability of the learned representations. It also includes a clear analysis of whether training in the 2D or 3D domain offers clear benefits for different applications.

It is interesting that deep learning methods are learning from essentially the same energy functions as classical methods using similar optimization approaches (e.g.,~stochastic gradient descent). The difference is that backpropagation updates are averaged over batches and whole datasets, seemingly alleviating problems of local minima or overfitting to a single sample. The problem then becomes overfitting to the \emph{distribution} of faces in the training set.
The training data used in these learning-based methods are often biased (e.g.,~\citet{liu2015faceattributes} includes mostly smiling faces). This leads to biases in the reconstruction methods. 
A practical question that requires to be solved is to determine the minimum amount of data required to apply deep learning methods. This is important when high-quality data is used for supervised training.

As learning-based and analysis-by-synthesis methods come together through self-supervised reconstruction methods, there are many shared challenges such as perspective face shape ambiguities and dealing with occlusions (e.g.,~\citet{tran2018extreme} already did a first step in this direction).
Learning-based methods typically are very fast and robust to initialization but achieve lower quality results compared to analysis-by-synthesis methods. One way to combine the desirable properties of these different paradigms is to use the learning-based solution as initialization for analysis-by-synthesis optimization~\cite{Tewari2018}.

While some recent methods have tried to build 3DMMs just from 2D data for better generalization, the resulting models are not as high-quality and lack details due to the low resolution of faces in currently available in-the-wild images. Bridging the gap in terms of details between models trained using high-quality data, and those built using only 2D data is an important open challenge.

Other challenges include extending recently developed methods to new applications. For instance, while monocular face reconstruction has started being explored, there is not yet much work on reconstructing a coherently deforming facial geometry from 2D video data.



%


\begin{figure*}[t]
  \centering
  \includegraphics[width=\linewidth]{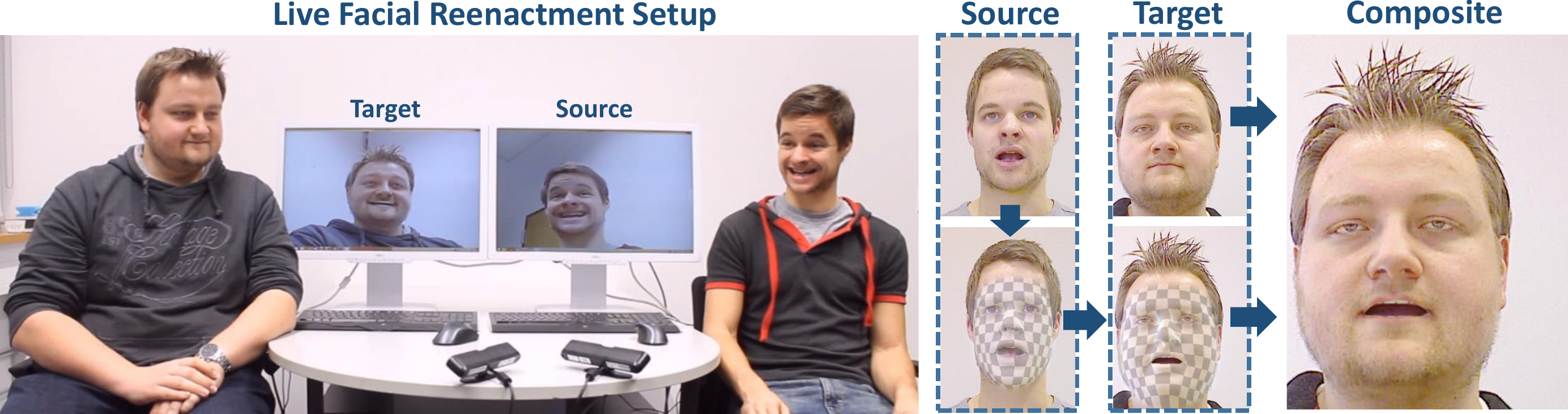}
  \vspace{-0.5cm}
  \caption
  {
  The first real-time facial reenactment approach \cite{Thies15} was based on RGB-D sensors.
  The approach tracks the facial expressions of a source and target actor, transfers the expression from source to target, and re-renders the target actor with the new expression on top of the input video stream.
  }
  \label{fig:thies15}
\end{figure*}

\section{Applications}
\label{sec:applications}
Parametric face models enable many compelling applications.
In the following, we will discuss applications in the domains of face recognition, entertainment, medical applications, forensics, cognitive science, neuroscience, and psychology.
All these applications have been pushed by the availability of publicly shared models and code (see Tab.~\ref{table:models}), as well as other resources \cite{OverviewGithub2019}.

\subsection{Face Recognition}
In the context of face recognition, 3DMMs have a manifold of potential applications.
\citet{blanz2002face} proposed to perform face recognition using the cosine angle on the shape and color coefficients estimated from a pair of 2D images as a distance metric for identification and recognition.
This distance metric exploits the natural disentanglement of 3DMMs separating identity (shape and color) from camera and illumination variation. It was shown that this 3DMM-based distance metric enables to recognize faces across large pose and illumination variations \cite{blanz2002face,blanz2003face,paysan20093d}, while being robust to facial expressions \cite{bfm17}, as well as being able to do recognition from features in the facial texture \cite{pierrard2007}.
Recently, \citet{tuan2017regressing} have shown that the performance of face recognition with 3DMM parameters can be enhanced by specifically taking the face recognition task into account when regressing the 3DMM parameters.
Whilst most work focuses on face recognition from 2D images, the 3DMM was also applied to the 3D face recognition task focusing on the shape coefficients and robust recognition with respect to facial expressions \cite{amberg2008expression,ter20083d,paysan20093d}.

Although 3DMMs have shown promising results at face recognition in controlled settings, they did not achieve a convincing performance on in the wild data. This arises from the ill-posed problem of estimating shape and color parameters from a 2D image,  at the same time a high precision of this estimation is needed for face recognition. Therefore, purely data-driven approaches have remained the dominant approach to face recognition, in particular since the advancement of deep learning technology \cite{taigman2014deepface,schroff2015facenet,Parkhi15}. 
However, data-driven approaches have fundamental problems such as their dependence on large-scale training data and their lack of generalization to out-of-distribution samples \cite{klare2012face}. One of the main issue for face recognition is the alignment of images. Careful alignment of face images has a big impact on face recognition accuracy and even for state of the art deep learning systems. 3DMMs are particularly useful for tackling these limitations, e.g., by using 3DMMs as a tool for face frontalization \cite{blanz2005face,tena20072d,hassner2015effective}.
In this context, it was shown that 9 out of 10 2D algorithms in the Face Recognition Vendor Test 2002 \cite{phillips2003face} improved considerably when combined with a 3DMM for face frontalization \cite{blanz2005face}.
Other applications of 3DMMs include augmenting real-world data in 3D \cite{masi2016we} and the generation of synthetic data for training \cite{SelaRK17, kortylewski2018training} and for analyzing the effects of dataset bias on face recognition systems \cite{kortylewski2018empirically,kortylewski2019analyzing}. 

Almost all applications of 3DMMs in the context of face recognition would benefit from improvements of the parametric model as well as the fitting process. A more realistic texture model including textural details and modeling hair would enhance the quality of synthetic data, possibly further reducing the amount of real-world data needed to train data-driven models. A more accurate fitting process would enhance the model's performance on face frontalization and face recognition from 3DMM parameters.

%

\subsection{Entertainment}
\label{sec:app:entertainment}
3DMMs are an integral building block for many compelling applications in the entertainment sector.
Such applications normally have to work in the wild and based on a low number of sensors, e.g., only the images captured by a single color camera are accessible.
In such underconstrained scenarios, the statistical prior that is encapsulated in the face model is a powerful tool to better constrain the underlying reconstruction problems.
In the following, we discuss several entertainment applications in detail.
These applications are also covered in more depth in the state of the art report of \citet{ZollhoeferSTAR2018}.

\subsubsection{Controlling 3D Avatars for Games and VR}
Realistic 3D face avatars can be reconstructed based on multi-view video \cite{Lombardi2018deep}, a few images \cite{Ichim2015,Cao16} or given only a single image \cite{Hu2017, wang2019digital}.
Such avatars or even artist-designed characters can be controlled in gaming scenarios based on dense trackers that employ a parametric face model.
Such vision-based control was first demonstrated in an off-line setting \cite{Chai03,Chuang02,Weise09,Wang04,Pighin06}.
Nowadays, dense facial performance capture is feasible at real-time rates based on RGB-D \cite{Weise11,Li13,Thies15} and color \cite{Bouaziz13,Thies16} cameras.
Besides vision-based animation, there is extended work on audio-based control \cite{VOCA2019,Karras:2017,TaylorKYMKGHM2017,Kshirsagar03}.
Face tracking can also be used to enable face-to-face communication \cite{Thies16VR,olszewski2016high,Lombardi2018deep,li2015facial} in virtual reality.

\subsubsection{Virtual Try-On and Make-Up}
Face reconstruction and tracking based on a parametric face model can also be employed to build virtual mirrors that enable the try-on of accessories or make-up.
To this end, first, a personalized model of the face is recovered and tracked across the video resulting in a dense set of correspondences.
These enable spatio-temporal re-texturing, e.g., to virtually place tattoos \cite{Garrido14} and can be used to add facial make-up \cite{bronstein2007calculus} and try out different suggestions \cite{Scherbaum:2011}.
Virtual make-up can be applied based on a reflectance/shading decomposition \cite{LiZhou14,LiZhou15}.
Similar techniques enable the try-on of accessories, e.g., eyeglasses \cite{Niswar2011,Azevedo2016}.

\subsubsection{Face Replacement a.k.a.~Face Swap}

Face replacement enables the replacement of the inner face region in a target video with that from a source video.
To this end, both persons are reconstructed based on the same parametric model resulting in dense inter-person correspondences.
First approaches enabled face replacement between images \cite{Blanz04,Jones08,Bitouk08,Kemelmacher-Shlizerman:2016}.
Later works extended those ideas including skin and hair segmentation to deal with glasses and occlusion by hair \cite{pierrard2008skin}.
Other techniques focus on swapping faces between video sequences \cite{Dale11,Garrido14}.
Today the effect is mostly known under the term `face swap' and has been popularized by a SnapChat\footnote{\url{https://www.snapchat.com/}} filter.

\subsubsection{Face Reenactment and Visual Dubbing}
\label{Deepfake}
Facial reenactment is the process of transferring the facial expressions from a source to a target video.
First, off-line techniques have been proposed \cite{Bregler97,blanz2003reanimating,Vlasic05,LiDai14,LiXu12,Kemelmacher10,Theobald09}.
The first real-time facial reenactment approach \cite{Thies15} was based on an RGB-D sensor, see Fig.~\ref{fig:thies15}.
Afterward, also real-time techniques for reenacting standard video have been proposed \cite{Thies16}.
Other approaches enable to take control of a single image \cite{Saragih11b,AverbuchCKC2017}.
Follow-up work focused on controlling more than just the face region, e.g., the complete upper body \cite{thies2018HeadOn}.
Nowadays, many reenactment approaches are based on deep generative models \cite{kim2018DeepVideo,pumarola2018ganimation}.

Facial reenactment \cite{kim2018DeepVideo,Thies16} can also be applied to the problem of visual dubbing, i.e., the task of adapting the mouth motion of a target actor to match a new audio track.
More sophisticated visual dubbing approaches \cite{Garrido15} directly take the new audio track into account for better audio-visual alignment. 
There is also some work on audio-based animation of video \cite{Brand99,Suwajanakorn:2017}.


\subsection{Medical Applications}
%
%
The clinical applications of the 3DMM cover both, analysis as well as synthesis. The dominant applications lie in analysis, where diseases can be recognized by facial shape. One example of such an effect is the classification and early diagnosis of fetal alcohol spectrum disorder \cite{suttie2013facial} or epilepsy \cite{ahmedt2019multi}. Similarly, \citet{hammond20043d} demonstrated both visualization and recognition of congenital craniofacial growth disorders. Both these works used 3D data. However, the capability of 3D reconstruction from 2D images was explored for the screening of acromegaly \cite{learned2006detecting} and genetic disorders \cite{tu2018analysis}.

In the direction of synthesis, the 3D shape model was explored to perform reconstruction of missing face parts based on the model statistics \cite{basso2005statistically,mueller2011missing}. Such a reconstruction can be applied for personalized implant design. Another work explored the synthesis capabilities for analysis and generated controlled stimuli to study responses in the fusiform face area and correlates them with autism spectrum disorder \cite{jiang2013quantitative}.

3DMM and statistical shape models, in general, are a popular standard framework in the field of medical imaging for segmentation and as models of variations in anatomical structures \cite{zheng2017statistical}. A lot of those applications deal with pathologies in young or elderly people which are underrepresented even in the biggest face models \cite{Ploumpis_2019_CVPR}. Those applications would profit from models built from a wider population or models than can better generalize beyond the data they are trained on.

\subsection{Forensics}
%
%
Applications in forensics range from identikit pictures over virtual aging to face reconstruction from dry skulls and recently also the detection of manipulated videos.

Describing faces from vague mental images is a challenging task. A tool based on a 3DMM \cite{blanz2006creating} allows exploring correlations within the face to generate indentikit pictures when providing descriptions based on vague features.

Virtual aging is a challenging task and can be helpful to later find missing children or victims of sexual abuse. The 3DMM helps to reduce the subjectivity of age progression methods. Several works in this direction are modeling age trajectories on 3DMM shape \cite{hutton2003estimating,koudelova2015modelling,shen20143d} and at least two attempts have been made to do so for both 3DMM shape and texture \cite{scherbaum2007prediction,hunter2009visual}. Most methods focus on children and neglect textural details or wrinkles which are modeled in \citet{paysan2010thesis,schneider2019}.

Face reconstruction from dry skulls is an ill-posed problem. The mapping from the skull to face is not a one-to-one, but a one-to-many mapping. Models allow to control attributes for this reconstruction \cite{paysan2009face}, explicitly estimate the posterior solution of possible faces per skull \cite{madsen2018probabilistic}
or model soft tissue thickness directly grounded by a 3DMM \cite{gietzen2019method}.

Recently 3DMMs were used successfully to generate or manipulate images and videos as discussed in Chapter~\ref{Deepfake}. At the same time 3DMMs are also helpful to detect those manipulations from state of the art methods with high accuracy \cite{rossler2019faceforensics++}.

\subsection{Cognitive Science, Neuroscience, and Psychology}
\label{sec:app:humanFaceProcessing}
%
%
The ability to generate faces that can be controlled via parameters is very popular when studying how the human and non-human primate brain process faces. Studies with generated stimuli from a 3DMM can be found in Cognitive Science, Neuroscience, Psychology, and Social Science.

One of the earliest works using 3DMMs presented high-level aftereffects that indicate a model related to a statistical face model in the human brain. Those aftereffects were demonstrated using caricaturized faces and antifaces \cite{leopold2001prototype}. Later it was shown that those results can not only be observed as aftereffects but also as responses of single neurons across caricaturization in macaque monkey to principal axes of a 3DMM \cite{leopold2006norm}. Later those aftereffects were shown to incorporate 3D information \cite{jiang2009three}. The effects based on caricatures for recognition were recently also investigated with 3DMMs in artificial neural networks trained on face recognition \cite{hill2018deep}.

A topic that was heavily researched over the past decades and is still under investigation is how much the 3D shape contributes to face perception and if the face representation in our brain is built as a 3D model. Early studies based on functional MRI and behavioral techniques evaluated a shape-based model of human face discrimination \cite{jiang2006evaluation}. Later studies investigated the importance of 3D shape and surface reflectance \cite{jiang2009neural} and event-related potentials to 3D shape are faster than to surface reflection \cite{caharel2009recognizing}. Other work explored how well humans can estimate a profile picture from a frontal view \cite{schumacher2012facial}.
    
Recently it was shown that a face-processing system based on stepwise inverse rendering correlates better to neural measurements in macaque monkey than state of the art artificial neural networks \cite{yildirim2019efficient}.

Face image manipulation is another key application of the 3DMM to generate stimuli \cite{walker2009portraits} to e.g., investigate social judgments based on facial appearance. Again the ability to control exactly what is manipulated is key for those research results sometimes measuring subtle effects \cite{walker2011universals}. Recently a dataset of controlled manipulated images was released to perform such experiments \cite{walker2018basel}. 

One of the major limitations compared to 2D based methods is that 3DMMs do not include hair. In a lot of studies faces and hair are not separated since faces without hair appear less face-like. For those models, it plays a substantial role to have controls over the parameters and that parameters can be interpreted which secures the future of 3DMMs in those fields.


\section{Perspective}
\label{sec:perspective}

In this last section, we want to look beyond the state of the art. We explicitly highlight the unsolved challenges in the field. In addition to focusing on face models, we look further and share our thoughts about the scalability of 3DMMs beyond faces. We also share our thoughts of the applicability of models including data, model and algorithm sharing also with its potential of misuse. We close with an outlook on how a 3DMM could look like in 10 or 20 years.

\subsection{Global Challenges}
In this section, we summarize the major open challenges that are shared across the different parts of 3DMMs. Local challenges that are specific to capturing, modeling, image formation or analysis-by-synthesis are mentioned in the respective sections.

One of the leading challenges is the balance between a low-dimensional parametric model and the degree of detail we are capable of modeling. Parametric models for eyes, teeth, hairs, skin details, soft tissue or even anatomical grounded muscles are not available. Additional complexity also renders analysis-by-synthesis even more challenging. Building faces with all those details is currently possible for a single face with a lot of manual labor, but automatic methods to extract those details or build models on top of them are in their beginnings. Current state of the art methods from capture, to modeling over image formation to analysis-by-synthesis use a lot of oversimplifying assumptions. Besides including more facial details there are also models that exploit the knowledge that a face is part of the body. Whilst faces and bodies are mostly analyzed separately, there exist first models that include faces and bodies jointly~\cite{Joo2018, SMPL-X:2019}. \citet{SMPL-X:2019} presented first results indicating that fitting the whole body is also beneficial for the quality measured in the face region only.

Another major challenge is the comparability of all the components of a 3DMM. Already the modeling itself can only be evaluated on specific tasks and different models have a different focus and might perform better on a specific task. For analysis-by-synthesis, comparing the performance of a model and also of the model adaptation algorithm is an unsolved problem. Current state of the art research frequently focuses on task-specific qualitative results and those results can barely be compared across models and algorithms. The current trend in the community to share source-code and models helps to compare and reproduce results, however, there is a lack of useful benchmarks. A first step in this direction is a new dataset providing natural images in combination with a 3D scan of the same individual \cite{RingNet:CVPR:2019}. However this is focused on shape reconstruction only, there is no single benchmark for 3D reconstruction from 2D images including illumination and albedo estimation. 

The last challenges are of an ethical nature. Concerns around image analysis and synthesis, especially for faces is currently discussed within the scientific community as well as in the media and the broad public. The current algorithmic development in computer vision and graphics allows to recognize faces and to generate or manipulate images and video. In addition most methods around 3DMMs elicit some dataset bias. \citet{saito2017photorealistic} approached this using the Chicago Face Databse\cite{ma2015chicago} to build a face model with balanced ethnicities. Those challenges are not a purely scientific one, but also a political one. We start to see regulations of those technologies and there will be likely more regulations across the world in the near future. As a community, we can choose on what projects we focus to work on and there are plenty of meaningful and valuable applications of 3DMMs, face analysis, and face synthesis as we presented in the Section~\ref{sec:applications} which could be explored less with restrictive regulations.

\subsection{Scalability}

Research on parametric models of human faces has seen a lot of progress in recent years. This raises the question of how scalable the found solutions are to other types of real-world entities beyond humans. 
On the one hand, human faces are highly challenging as we are attuned to noticing even slightest inaccuracies in their modeling. At the same time, they are also more amenable to statistical modeling as their structure is relatively regular and correspondence across faces is quite well-defined. Other types of real-world entities, or even humans in clothing or the human head with full hair, are exhibiting much stronger appearance, structure, and shape variation that may require additional methodical innovations to empower proper modeling. The vision and graphics communities have begun to build and learn statistical models of other types of shape categories. Researchers also increasingly attempt to learn such models in an unsupervised or weakly supervised way for better real-world scalability. These approaches partially build on many concepts learned from the models described in this article but introduce additional representation innovations, like learned implicit representations \cite{cole2017synthesizing,eslami2018neural, sitzmann2019srns}, to handle their specific structural properties. Future research will certainly see more work in this direction that answers the question of what is the right shape, appearance and deformation representations for a wider range of real-world object classes. 

\subsection{Application}

An additional challenge for our research community will be to agree on efficient ways to share and combine research efforts performed by different research groups. We should agree on common data formats and dissemination channels for available scan databases, which would simplify building integrated models, and enable us to better test and compare them. 
In that context, ever more pressing questions of privacy and security will also need to be addressed. On the one hand, it is needless to say that we have to adhere to highest standards of privacy protection in data sets we share, so not to reveal personal data or identities beyond what is needed and permitted by law or by the captured individuals. For handling this, community-wide procedures for providing consent on the use of data that are compatible with legal regulations could be agreed on and shared.

However, beyond this, increasingly powerful methods to build and reconstruct such face models from image and video will in the future enable us to build highly believable 3D human avatars from casually captured imagery. These avatars will enable us to create virtual renditions of real people at unseen accuracy to populate computer graphically generated virtual spaces at high visual fidelity. 
However, algorithmic tools should be investigated as well that prevent the reconstruction or use of such avatars in undesired or questionable applications that a reconstructed person did not provide consent on.  
Advanced reconstruction algorithms on the basis of parametric models may also make it possible to extract semantic information of people from imagery that they may not want to reveal (e.g., about emotional state, health, and physical condition, etc.). Therefore, algorithmic strategies to balance personal privacy and reconstruction ability shall be investigated and provided by our research community. 

Also, the continuously improving performance of algorithms to reconstruct detailed human models from single images or videos enables advanced new ways to synthesize new face imagery or even modify existing face images and videos at very high visual fidelity. As an example, some recent combinations of model-based reconstruction algorithms and adversarially trained neural networks have shown impressive results in that respect. Such advanced synthesis algorithms will simplify many applications and open up entirely new applications, for instance in content creation for animation and visual effects, in content creation for virtual and augmented reality, in telepresence, visual dubbing or advanced video editing. 
However, they might also be used to create or modify media content with malicious intent. 
Therefore, as a community focusing on basic research, we will continue our efforts to objectively inform the general public about the great possibilities opened up by advanced parametric models of face, body and other real-world entities to build the next generation of intelligent, interactive and creative computing systems. At the same time, we will use our essential basic expertise about the underlying algorithmic principles to develop new ways to detect unwanted media synthesis and modification and to prevent such unwanted modifications algorithmically. 

\subsection{Outlook}
The big question we ask is how will a generative face model look like in 10 or 20 years? What will be the representation and will it be a complete model of the human face with all its variation and details? Currently, we experience a divergence of 3DMMs. Different research teams put a different focus and model some parts in more detail but lack other details or statistical variation. Recent modeling advances are focusing on building task-specific representations rather than a more general face model to be applicable for multiple tasks. For some applications, the model itself is the limiting factor, whilst other applications profit from a simple model based on PCA. The requirements in terms of quality, realism, generalization, and performance are very different e.g., content creation vs. computer vision. The gap between state of the art computer-generated renderings for a single face including expressions versus generative and parametric face models based on statistics is dramatic.

Current advances in the field of machine learning will contribute to build more general and at the same time more realistic models. The core of the face model was always interpreted as a learning problem, recent advances lifted the analysis-by-synthesis task from a per image optimization task to a learning challenge. However, this loop is not yet closed - why not learn or improve the model itself? There are already first works in the direction of model learning (compare Section~\ref{sec:3Dfrom2D}), but they are limited by very similar modeling assumptions as traditional 3DMMs. First steps to overcome those were recently performed in the direction of neural rendering \cite{eslami2018neural,thies2019deferred}, 3D representation learning \cite{sitzmann2019srns} and unsupervised shape model learning \cite{szabo2019unsupervised}. Other modeling approaches like generative adversarial networks \cite{goodfellow2014generative,Karras2019cvpr,Karras2019stylegan2} are currently operating in 2D image space. Such parametric models can be used to embed faces of real people in a latent space \citet{Abdal_2019_ICCV,abdal2019image2stylegan}, but the resulting embedding is hard for humans to interpret.

20 years ago 3DMMs were part of a revolution in computer graphics and computer vision to go away from 2D image processing to 3D modeling. The computer vision community is currently focusing again on mainly 2D based approaches and we have to propose the missing key to again move the community to 3D. 
Additionally one of the leading benefits of 3DMM is the natural disentanglement of shape, color, illumination and camera parameters. Such a disentanglement is very hard to be derived purely from data \cite{locatello2018challenging} and for faces, 3DMMs build it manually based on the image formation process.
According to "Pattern Theory" \cite{ grenander1996elements, mumford2010pattern} it is a prerequisite for any high-performance image analysis system to find and separate conditional independent parameters that describe the image to analyze. The discovery and separation of such parameters purely from 2D data is still an unsolved challenge. 3DMMs directly implement models using the parameter also used by physics and geometry to model light and three-dimensional objects.  

One direction which might be particularly interesting is to break out of the common modeling assumptions and oversimplification but at the same time automate the tedious manual work behind the photo-realistic generation of faces. We expect some kind of living 3DMM to evolve from the community. Automation will be the leading modeling idea. A living 3DMM should be able to learn from 3D data as well as 2D data, both still and in motion. We imagine the model to be learned from a minimal seed like a mean face, a sphere or just a rough prototype based on the first few data points. The optimal living model would not be task-specific but should be able to generalize to various tasks. The face model must, therefore, be hierarchical in some form to represent multiple degrees of detail but share statistics across those levels. Such an optimal face model would be general enough to be applicable for real-time computer vision tasks, analysis-by-synthesis from currently challenging images as well as photorealistic rendering with a high level of facial details. Last but not least some tasks rely on an interpretable parametrization and not just a black box learning machine. Basic knowledge of geometry and physics would not only ease the learning but also at least disentangle pose and illumination variation from the facial shape and appearance. Building such a general face model might remain a challenge for the next 10 or 20 years but would align with the original idea behind 3DMMs.

\begin{acks}
This survey paper was initiated at the Dagstuhl Seminar 19102 on 3D Morphable Models \cite{dagstuhl2019momo} and contains ideas resulting from discussions at this seminar.
This survey paper was partially funded by Early PostDoc Mobility Grant, Swiss National Science Foundation P2BSP2\_178643, ERC Consolidator Grant 4DRepLy and the Max Planck Center for Visual Computing and Communications (MPC-VCC). We thank Bar\i{}\c{s} Ge\c{c}er for his help on the teaser figure, and Haiwen Feng for providing the FLAME texture space. We thank the anonymous reviewers whose comments have greatly improved this manuscript.
\end{acks}

\bibliographystyle{ACM-Reference-Format}
\bibliography{20yMoMo}

\end{document}